%% file: arxiv.tex
\pdfoutput=1
\documentclass[10pt,twocolumn,letterpaper]{article}
\def\arxivversion{}
\usepackage[pagenumbers]{cvpr}
\input{preamble}

\definecolor{cvprblue}{rgb}{0.21,0.49,0.74}
\usepackage[pagebackref,breaklinks,colorlinks,allcolors=cvprblue]{hyperref}
\hypersetup{
  pdftitle={Seeing Is Not Planning: Spatiotemporal Planning and Plan Repair in Multimodal LLMs},
  pdfauthor={Siyuan Li, Ethan Lin, Xiang Li, Tianming Liu}
}

\def\confName{CVPR}
\def\confYear{2027}

\title{Seeing Is Not Planning:\\Spatiotemporal Planning and Plan Repair in Multimodal LLMs}
\author{
Siyuan Li$^{1,2}$\qquad Ethan Lin$^{1}$\qquad Xiang Li$^{2}$\qquad Tianming Liu$^{1}$\\[4pt]
$^{1}$University of Georgia\qquad $^{2}$Massachusetts General Hospital, Harvard University
}

\begin{document}

\twocolumn[{%
\renewcommand\twocolumn[1][]{#1}%
\maketitle
\vspace{-1.2em}
\begin{center}
\includegraphics[width=\textwidth]{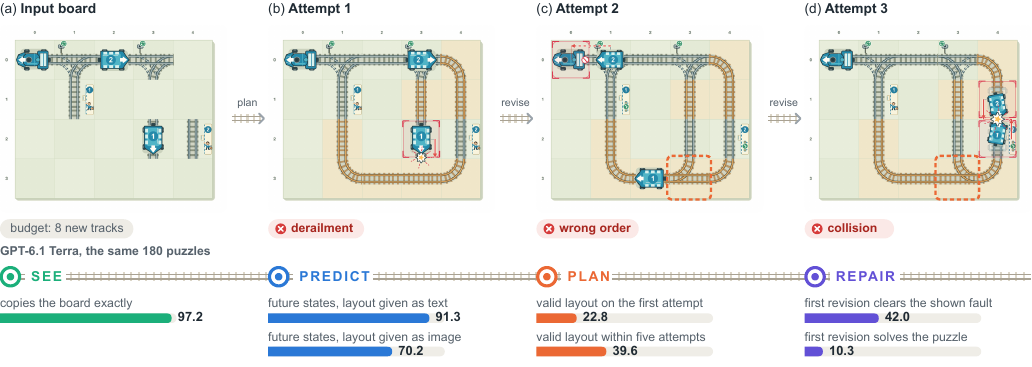}
\vspace{-1.2em}
\captionof{figure}{\textbf{Spatiotemporal planning and plan repair from visual input.}
\emph{Top:} the model adds tracks to the board in (a); both cars then move at once, and each must stop at its own station (the numbered platform) before car~1 and then car~2 reach the engine.
Every attempt fails. Each revision changes only the dashed cell, which removes the previous failure and creates a new one elsewhere.
\emph{Bottom:} diagnostics for \terra on the same 180 puzzles (\%). Reading the board and predicting how a given layout runs are much easier than building a layout that works. Repair is scored on the first revision of 417 failed answers: the revision often clears the fault it was shown, but rarely solves the puzzle.}
\label{fig:teaser}
\vspace{0.8em}
\end{center}%
}]

\input{sec/0_abstract}
\input{sec/1_intro}
\input{sec/2_related}
\input{sec/3_benchmark}
\input{sec/4_results}
\input{sec/5_analysis}
\input{sec/6_conclusion}

{
\small
\bibliographystyle{ieeenat_fullname}
\bibliography{main}
}

\clearpage
\appendix
\twocolumn[{\centering\Large\textbf{Appendix}\par\vspace{1.0em}}]
\input{sec/supp}

\end{document}

%% file: preamble.tex
\newif\ifarxiv
\ifdefined\arxivversion\arxivtrue\fi
\usepackage{booktabs}
\usepackage{array}
\usepackage{tikz}
\usepackage{xspace}
\usepackage{caption}
\usepackage{amsmath}
\usepackage{amssymb}
\usepackage{microtype}

\newcommand{\bench}{Railbound-Bench\xspace}
\newcommand{\sol}{GPT-6.1 Sol\xspace}
\newcommand{\astra}{GPT-6 Astra\xspace}
\newcommand{\terra}{GPT-6.1 Terra\xspace}
\newcommand{\luna}{GPT-6 Luna\xspace}
\newcommand{\haiku}{Claude Haiku 5.5\xspace}
\newcommand{\deepseek}{DeepSeek-V4.1 Flash\xspace}
\newcommand{\minimax}{MiniMax-M3.1 Flash\xspace}
\newcommand{\qwenbig}{Qwen3.5-397B-A17B\xspace}
\newcommand{\qwen}{Qwen3.5-35B-A3B\xspace}
\newcommand{\Sk}[1]{S@#1}
\newcommand{\elem}[1]{\raisebox{-0.12em}{\tikz\node[circle,fill=black!85,text=white,inner sep=0pt,minimum size=1.15em,font=\sffamily\bfseries\scriptsize]{#1};}}

\ifarxiv\else
\fi
\ifarxiv
\newcommand{\suppref}[2]{\cref{#1}}
\newcommand{\mainref}[2]{\cref{#1}}
\else
\newcommand{\suppref}[2]{Supp.~#2}
\newcommand{\mainref}[2]{#2}
\fi
\newcommand{\codeurl}{\url{https://github.com/liaoyanqing666/Railbound_Bench}}
\newcommand{\tablestyle}{\footnotesize\setlength{\tabcolsep}{4pt}\renewcommand{\arraystretch}{1.05}}

%% file: sec/0_abstract.tex
\begin{abstract}
Planning in a visual world means reading the layout, foreseeing how the state will change, and following the task rules all at once.
When a plan fails, a model should also use what went wrong to fix it.
We call these two abilities spatiotemporal planning and plan repair, and test them with \bench.
A model sees a railway puzzle, adds tracks under a budget, and submits one complete layout; a simulator then moves all cars at once and decides whether the layout works.
After a failure, the model sees what happened and may submit a new layout, up to five attempts in total.
On the 200-puzzle main set, the best of nine models solves 59.5\% within five attempts, the next 39.5\%, and the other seven at most 10.0\%; a 1,000-puzzle extended set with a wider range of difficulty is far easier, at 90.9\% for the best model.
Controlled studies on 180 fixed puzzles show that knowing the task is not the same as using it.
\terra copies 97.2\% of boards exactly and, given a correct layout as text, predicts all four requested future states in 91.3\% of cases, yet only 22.8\% of its first layouts work.
Planning fails more often as the cars' routes interact, even among puzzles that a search solver finds equally easy.
Visual feedback removes the reported fault in 175 of 417 first revisions, but only 43 of these revisions solve the puzzle, and at an equal budget of 40k tokens it does no better than fresh retries.
\ifarxiv The code and data are open-sourced on GitHub at \codeurl.\fi
\end{abstract}

%% file: sec/1_intro.tex
\section{Introduction}
\label{sec:intro}

Many plans are established before any action takes place.
A railway timetable, a warehouse floor plan and a traffic-light schedule must all work once the trains, robots or cars start moving, and the planner cannot intervene.
Making such a plan requires reading the layout, foreseeing how the state will change over time, and keeping every rule satisfied at once.
We call this ability \emph{spatiotemporal planning}.
When a plan fails, a second ability matters: using what went wrong to produce a better plan.
We call this \emph{plan repair}.

People often predict how a physical system will behave by mental simulation, running a model of it forward in their heads~\cite{hegarty2004mechanical,battaglia2013simulation}.
Yet reading a scene, predicting how a given plan will run, building a plan, and repairing a failed plan are different tasks.
A model may answer questions about a board correctly and still fail to choose moves that work together.
A revision may fix the reported error and still leave the plan broken somewhere else.
Measuring these abilities on the same puzzles shows where knowing the task stops turning into working plans.

Existing benchmarks test visual puzzles~\cite{chia2024puzzlevqa,song2025visualpuzzles}, mental simulation of visual changes~\cite{li2026stare}, and planning in visual environments~\cite{wu2024vsp,paglieri2025balrog}.
Recent work also studies spatiotemporal reasoning and plan revision from feedback~\cite{lai2026ustbench,shi2026spaceact,tupini2026asgardbench}.
We focus on one comparison that these works leave open: building a complete plan and repairing it after it has run, within one family of executable tasks, together with matched tests of perception, prediction, rule knowledge, and feedback.
Unlike interactive agents, which can change course during a run, our model submits the whole plan first and sees feedback only after the plan has run.

We build this evaluation on the rules of the puzzle game Railbound~\cite{railbound} (\cref{fig:teaser}).
The model sees an image of a board with rails, cars, an engine, and mechanisms such as gates, tunnels, and stations.
It adds track pieces under a budget and submits one complete layout.
After receiving the model's answer, our simulator runs the cars, moving each car one cell along the track per step, with no further input from the model.
The layout works only if every car stays on connected rails, stops where it must, avoids the other cars, and reaches the engine in its turn.
If it fails, the model receives a short message, a text description, or images of the failed run, and may submit a new complete layout, up to five attempts in total.
The simulator checks every answer exactly, and a search solver provides reference solutions and a measure of how much search each puzzle needs.

The main leaderboard uses 200 hard, newly designed puzzles (Hard200); a further 1,000 puzzles (Extended1000) mix easy and hard ones.
On Hard200, the best of nine models solves 59.5\% within five attempts, and seven solve at most 10\%.
Controlled studies on a fixed panel of 180 Extended1000 puzzles link planning and repair to the skills that support them, and lead to three findings:
\begin{itemize}
\item \textbf{Understanding the task does not mean being able to plan it.}
\terra copies 97.2\% of boards exactly and, given a correct layout as text, predicts all four requested future states in 91.3\% of cases.
Yet only 22.8\% of its first layouts work, and giving it the exact board as text helps only modestly.
\item \textbf{Knowing how mechanisms behave and combining constraints are different problems.}
Examples of a mechanism on other boards teach the model how it behaves, but planning still breaks down as the cars' routes interact, even after accounting for solver search effort and other puzzle properties.
\item \textbf{Fixing the reported error is not the same as repairing the plan.}
Visual feedback removes the observed fault 2.5 times as often as a fresh retry, yet 132 of the 175 revisions that remove it fail somewhere else.
At 40k tokens, visual feedback is no better than fresh retries.
\end{itemize}

Our contributions are:
(i) an executable benchmark for visual spatiotemporal planning and plan repair, with 1,200 newly designed puzzles, a deterministic simulator and a tool-free protocol;
(ii) controlled diagnostics that link perception, prediction, rule knowledge, plan construction, and repair on the same puzzles; and
(iii) an evaluation of nine models from five providers that separates first-attempt planning, recovery after failure, and cost.

%% file: sec/2_related.tex
\section{Related Work}
\label{sec:related}

\begin{figure*}[t]
\centering
\includegraphics[width=\textwidth]{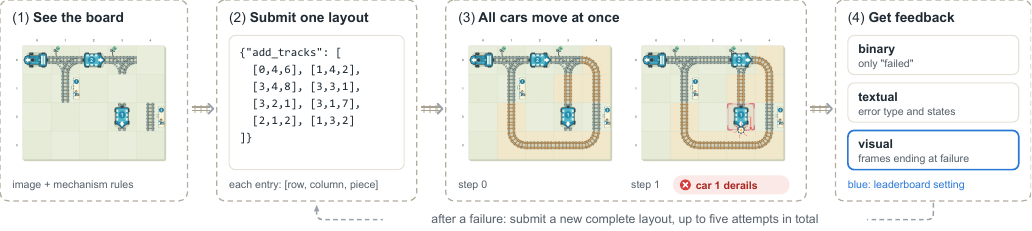}
\caption{\textbf{Task and protocol.} (1) The model receives the board image and the rules of the mechanisms on it. (2) It returns all the tracks it adds as one list. (3) The simulator moves all cars at once; here, car~1 derails at the first step. (4) After a failure, the model receives one kind of feedback and submits a new complete layout, up to five attempts in total. The leaderboard uses visual feedback.}
\label{fig:task}
\end{figure*}

\paragraph{Spatiotemporal reasoning and planning.}
Visual puzzle benchmarks test pattern reasoning and constraint satisfaction~\cite{chia2024puzzlevqa,ren2025vgrpbench}, STARE tests whether models can mentally simulate visual changes~\cite{li2026stare}, and VSP separates perception errors from reasoning errors in spatial planning~\cite{wu2024vsp}.
Recent benchmarks target spatiotemporal abilities more directly.
USTBench splits urban reasoning into understanding, forecasting, planning, and reflection~\cite{lai2026ustbench}.
ET-Plan-Bench varies spatial constraints and time dependencies in embodied tasks~\cite{zhang2025etplanbench}, and SpaceAct tests 3D geometry, physical dynamics, and action refinement in interactive puzzles~\cite{shi2026spaceact}.
Like these works, we treat spatiotemporal planning as an ability built from several parts.
Our controlled studies measure those parts---reading the scene, predicting what happens next, and knowing the rules---on the same puzzles on which we measure planning and repair.

\paragraph{Prediction and executable plans.}
Predicting how a given system will evolve~\cite{wang2024simulators,gu2024cruxeval} is a different task from choosing a plan that reaches a goal.
Game benchmarks such as BALROG and VS-Bench test planning through repeated actions and observations~\cite{paglieri2025balrog,xu2026vsbench}.
In PHYRE, a player places objects once and physics decides the outcome~\cite{bakhtin2019phyre}; BuildArena and BesiegeField judge engineering structures and machine designs by running them~\cite{xia2026buildarena,zhang2026besiegefield}.
These works show that a static design can encode future behavior.
\bench applies this idea to planning from images under explicit spatial and timing rules, adds separate prediction tests, and judges first and revised plans by the same run.

\paragraph{Plan repair and adaptation.}
AsgardBench tests whether visual observations and minimal feedback help models adapt a plan during execution, with diagnostics of grounding and state tracking~\cite{tupini2026asgardbench}.
VLA-FixBench links fault diagnosis to rollback, spatial correction, and recovery in robot manipulation~\cite{yan2026vlafixbench}.
Both suggest measuring successful recovery separately from failure recognition.
We study repair between complete runs: the model sees evidence about a failed plan and submits a revised plan under the same rules.
We then check separately whether the revision removes the reported fault and whether the revised plan succeeds.

\paragraph{Feedback and compute budget.}
Whether self-correction helps depends on external feedback and on how well the model uses it~\cite{huang2024cannot,kamoi2024survey,jiang2025friction}.
In code generation, independent sampling can match self-repair at equal cost~\cite{olausson2024selfrepair,zheng2025multiturn}, and the value of step-by-step refinement depends on task difficulty~\cite{snell2025scaling}.
We compare feedback formats, follow revisions of the same failed answers, and compare feedback with fresh retries at equal token budgets.
As in planning studies with exact checkers~\cite{valmeekam2023planbench,kambhampati2024llmmodulo}, we use a solver only to verify solutions and measure search effort; the model handles all planning and repair.

%% file: sec/3_benchmark.tex
\section{Benchmark and Evaluation}
\label{sec:benchmark}

\begin{figure*}[t]
\centering
\begin{minipage}[c]{0.37\textwidth}
\includegraphics[width=\linewidth]{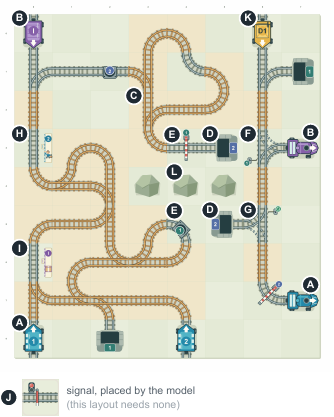}
\end{minipage}\hfill
\begin{minipage}[c]{0.605\textwidth}
\footnotesize
\renewcommand{\arraystretch}{1.25}
\begin{tabular}{@{}c@{\hspace{4pt}}p{0.44\linewidth}@{\hspace{10pt}}c@{\hspace{4pt}}p{0.44\linewidth}@{}}
\elem{A} & \textbf{Car and engine.} Every car moves one cell per step in the direction it faces. Cars must reach the engine in numerical order; a delivered car leaves the board. &
\elem{G} & \textbf{Self-switching points.} A three-way rail whose curve flips each time a car enters it. \\
\elem{B} & \textbf{Freight car.} Freight cars (Roman numerals) go to the freight engine, in their own number order. &
\elem{H} & \textbf{Passenger station.} The car with the matching number must stop here for two extra steps before it is delivered; other cars pass by. \\
\elem{C} & \textbf{Three-way rail.} A car entering at either end of the curve follows the curve; a car entering at the third end goes straight. &
\elem{I} & \textbf{Post office.} Works like a station, for the freight car with the matching numeral. \\
\elem{D} & \textbf{Tunnel.} Two portals with the same number are linked; a car entering one leaves from the other in the same step. &
\elem{J} & \textbf{Signal.} Placed by the model next to a three-way rail. It holds cars until another car passes that rail moving away from it, then stays open. \\
\elem{E} & \textbf{Button and gate.} A car crossing a button opens or closes every gate with the same number. Cars wait at a closed gate. &
\elem{K} & \textbf{Decoy car.} Need not be delivered and may not enter an engine. It can press buttons; if it derails, its wreck blocks the track. \\
\elem{F} & \textbf{Button-switched points.} A three-way rail whose curve flips when a car crosses the button with the same number. &
\elem{L} & \textbf{Rock or building.} Blocks travel; no track can be placed on it. \\
\end{tabular}
\end{minipage}
\caption{\textbf{Board elements and their rules}, shown on Extended1000 puzzle E-TR1501 with its reference layout. Grey rails are given; orange rails are the 30 pieces added in the reference layout. Letters mark one example of each element. Numbers printed on the board belong to the game: a shared number links objects that work together, such as a car and its station, or a button and its gates.}
\label{fig:elements}
\end{figure*}

\subsection{Task}
\label{sec:task}

\paragraph{Board and answer.}
A puzzle is a grid of cells (\cref{fig:task}).
Cells may hold fixed rails, rocks or station buildings, numbered cars, one or two engines, and mechanisms such as gates, tunnels and stations; \cref{fig:elements} lists every element and what it does.
Each car faces the direction in which it will move next.
The model may add straight, curved, or three-way track pieces to empty cells, up to a track budget, and in some puzzles a limited number of signals.
It returns all additions at once as a list of \texttt{[row, column, piece]} entries; this list is one complete layout.

\paragraph{How a layout runs.}
The simulator then runs the completed board in discrete steps.
At every step, all cars move one cell at the same time, unless a gate, a signal, or a station makes a car wait.
A car must reach an engine of its own type, and cars of each type must arrive in numerical order.
The run fails if two cars enter the same cell or swap cells head-on, if a car runs onto a missing or mismatched rail (a derailment), if a car arrives out of order or skips a stop, or if the board returns to an earlier state, which would repeat forever.
The run succeeds when every required car has been delivered.
Because all cars share the same rails and mechanisms, a piece that routes one car correctly can delay, block, or redirect another, so a layout can only be judged as a whole.
These rules test spatial construction, prediction of changing states, timing and cause-and-effect dependencies, limited resources, and coordination between cars.
Plan repair uses the same task and the same success test: after a failure, the model submits a new complete layout.

\subsection{Puzzles}
\label{sec:puzzles}

\paragraph{Search solver.}
We build and grade the puzzles with an exact search solver written for this benchmark.
It runs the simulator and, whenever a car is about to enter an empty cell, branches over the pieces that could be placed there; it drops a branch as soon as a move fails, the budget runs out, or a car can no longer reach its stops and engine.
Each choice it explores is one \emph{search node}.
Started without a reference solution, the number of nodes it explores before finding a first solution measures how much search a puzzle needs.
The same solver also searches for the fewest added tracks a puzzle needs, which sets the track budget.

\paragraph{Main set: Hard200.}
The main set contains 200 newly designed puzzles, each needing more than $10^5$ search nodes.
Its boards are larger, its budgets higher, and its runs longer than in the extended set (\cref{tab:stats}).
Every puzzle has a reference solution checked by the simulator, and its budget equals the fewest added tracks we found; we proved this minimum for 136 puzzles.

\paragraph{Extended set: Extended1000.}
A separately designed set of 1,000 puzzles uses the same rules over a wider range of difficulty.
It covers plain routing, delivery order, passenger and freight cars, tunnels, switching, gates, decoys and signals.
We designed it so that its difficulty mix, board sizes, budgets, and the share of puzzles with each mechanism stay close to those of the original game's 233 levels (\suppref{sup:puzzles}{C}).
By search effort, 598 puzzles are easy ($\leq10^3$ nodes), 300 medium ($10^3$--$10^5$) and 102 hard ($>10^5$), against 136, 73 and 24 of the original levels.
The solver solves 992 puzzles within 600 seconds, and the minimum budget is proven for 987; the remaining budgets come from the best valid solutions found within the search time.
These labels describe search effort, not how hard a puzzle is for a model.

\paragraph{Original levels and rule differences.}
We also use the 233 levels of the original game, as transcribed by the community~\cite{railboundsolver}, to check our simulator against known solutions; they keep their original track allowances and are not part of either leaderboard.
Our simulator was written independently and follows the rules of Railbound~\cite{railbound} with two changes: engines may stand inside the board, and a delivered car leaves at the end of its move without pausing the other cars.
All results follow our simulator's rules.

\begin{table}[t]
\caption{\textbf{Puzzle statistics}, as minimum--maximum (median). Run lengths come from the reference solutions. The original levels keep their own track allowances and serve only to check the rules.}
\label{tab:stats}
\centering
\tablestyle\setlength{\tabcolsep}{3.5pt}
\begin{tabular}{@{}lccc@{}}
\toprule
& Hard200 & Extended1000 & Original \\
\midrule
Puzzles & 200 & 1,000 & 233 \\
Board cells & 24--182 (50.5) & 6--108 (30) & 5--99 (32) \\
Cars, incl.\ decoys & 1--10 (4) & 1--6 (3) & 1--5 (2) \\
Track budget & 9--62 (19) & 1--43 (11) & 0--48 (11) \\
Run length (steps) & 12--439 (43) & 3--153 (17) & 4--188 (18) \\
Easy / medium / hard & 0 / 0 / 200 & 598 / 300 / 102 & 136 / 73 / 24 \\
Minimum budget proved & 136 (68.0\%) & 987 (98.7\%) & -- \\
\bottomrule
\end{tabular}
\end{table}

\paragraph{Validity.}
In every reference layout of the 1,200 designed puzzles, each rail branch connects to a matching rail, except the unused rear end of a starting car.
We reviewed every puzzle image and its solved image, removed near-duplicates that differ only in empty space or decoration, and kept no puzzle with the same gameplay in two sets.
Guessing does not work: across 200 random layouts per Extended1000 puzzle, only 0.01\% are valid.

\paragraph{Controlled panel.}
For the diagnostic studies we fix 180 Extended1000 puzzles (60 easy, 59 medium, 61 hard), chosen by puzzle properties to cover all mechanisms and board sizes, without looking at model results.
The panel is for controlled comparisons, not for estimating full-set accuracy.
We study \terra because it solves 50.0\% of Extended1000, the panel's parent set, within five answers: far from floor and ceiling, so an intervention can visibly raise or lower success, and cheap enough to run every condition three times.
We pool the three runs; conditions with the same first input share the same first answers, so they differ only in what happens after the first failure.
Confidence intervals (95\%) come from resampling puzzles (bootstrap).

\subsection{Protocol and metrics}
\label{sec:protocol}

Models answer without tools, code execution, or web access.
The prompt is in English and does not name the game.
It contains the board image, a legend explaining how pieces are drawn, the goal, the answer format, and the rules of the mechanisms on the board.
After a failed answer, the conversation keeps the puzzle, the earlier answers, and the feedback, but not earlier reasoning.
The leaderboard allows five graded answers with visual feedback: up to three rendered frames ending at the failure, four steps apart.
Each answer may use up to the provider's output limit, from 65,536 tokens to the model maximum (\suppref{sup:settings}{A}); no graded answer in any run reached it.
Malformed or illegal answers use up an attempt; service failures and empty replies do not.

We report \Sk{k}, the share of puzzles solved within $k$ graded answers, and the repair rate, the share of a model's first-answer failures that it solves later:
\begin{equation}
\mathrm{Rep.} = \frac{N_5 - N_1}{N - N_1},
\label{eq:repair}
\end{equation}
where $N_k$ is the number of puzzles solved within $k$ answers and $N$ the number of puzzles.

%% file: sec/4_results.tex
\section{Main Results}
\label{sec:results}

\paragraph{Models.}
We evaluate nine models from five providers: \sol, \astra, \terra and \luna from OpenAI, \haiku from Anthropic, \deepseek, \minimax, \qwenbig, and \qwen.
All reason before answering and run on Hard200; the five with entries in both columns of \cref{tab:leaderboard} also run on Extended1000.
Each model is run once per set; three repeated \terra runs of the panel's base condition differ by at most 2.7 points in \Sk{1} and 2.2 in \Sk{5}, and \suppref{sup:results}{B} gives 95\% bootstrap intervals for every entry.

\begin{table}[t]
\caption{\textbf{Planning and repair leaderboards} (\%). \Sk{k}: puzzles solved within $k$ graded answers. Rep.: share of first-answer failures solved by the fifth answer. Dashes: not run on Extended1000.}
\label{tab:leaderboard}
\centering
\tablestyle\setlength{\tabcolsep}{3.0pt}
\begin{tabular}{@{}l ccc ccc@{}}
\toprule
& \multicolumn{3}{c}{Hard200 (main)} & \multicolumn{3}{c}{Extended1000} \\
\cmidrule(lr){2-4}\cmidrule(l){5-7}
Model & \Sk{1} & \Sk{5} & Rep. & \Sk{1} & \Sk{5} & Rep. \\
\midrule
\sol & \textbf{30.5} & \textbf{59.5} & \textbf{41.7} & \textbf{66.8} & \textbf{90.9} & \textbf{72.6} \\
\astra & 19.5 & 39.5 & 24.8 & 59.6 & 83.5 & 59.2 \\
\deepseek & 2.5 & 10.0 & 7.7 & -- & -- & -- \\
\terra & 0.5 & 5.5 & 5.0 & 31.9 & 50.0 & 26.6 \\
\minimax & 1.0 & 4.0 & 3.0 & -- & -- & -- \\
\haiku & 2.0 & 3.5 & 1.5 & 26.6 & 50.9 & 33.1 \\
\luna & 0.5 & 2.5 & 2.0 & 7.5 & 16.1 & 9.3 \\
\qwenbig & 0.0 & 0.0 & 0.0 & -- & -- & -- \\
\qwen & 0.0 & 0.0 & 0.0 & -- & -- & -- \\
\bottomrule
\end{tabular}
\end{table}

\begin{figure*}[t]
\centering
\includegraphics[width=\textwidth]{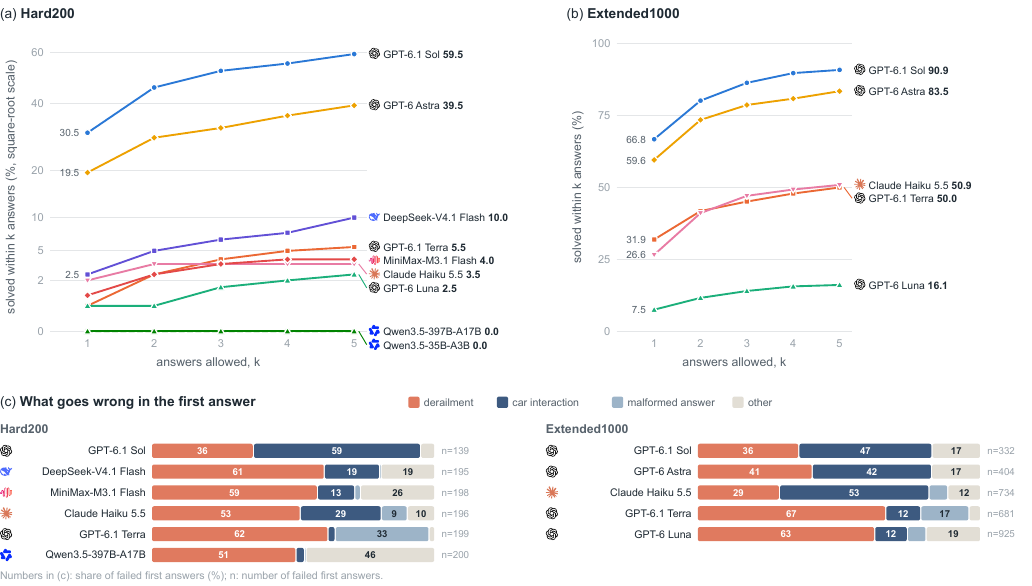}
\caption{\textbf{Leaderboards and first-answer failures.} (a, b) Puzzles solved within $k$ answers on Hard200, the main set, and on Extended1000; every answer after the first follows visual feedback on the previous failure. Panel (a) uses a square-root scale so that the low scores stay apart. (c) What the first answer gets wrong, as a share of failed first answers: derailment, car interaction (collision, wrong delivery order, or wrong engine), malformed answer (output that cannot be parsed or has the wrong structure), and other errors (illegal or over-budget edits, loops, missed stops). Each model keeps its color and marker across panels.}
\label{fig:curves}
\end{figure*}

\paragraph{Hard200 leaves much room.}
\Cref{tab:leaderboard,fig:curves} separate first-attempt planning from repair after feedback.
On Hard200, \sol solves 30.5\% on the first attempt and 59.5\% within five; \astra reaches 19.5\% and 39.5\%.
The other seven models solve at most 10.0\% (\deepseek), and both Qwen models solve none.
Extended1000 is much easier: \sol and \astra reach 90.9\% and 83.5\%, and their \Sk{5} falls from 98.2\% and 94.5\% on easy puzzles to 55.9\% and 46.1\% on hard ones (\suppref{sup:results}{B}).

\paragraph{First attempts and repair must be measured separately.}
On Hard200, \sol repairs 58 of its 139 first-answer failures (41.7\%), and \astra 40 of 161 (24.8\%); every other model repairs at most 7.7\%.
On Extended1000, \haiku starts below \terra (26.6\% versus 31.9\%) but ends slightly above it after feedback (50.9\% versus 50.0\%); their repair rates are 33.1\% and 26.6\%.
First-attempt success alone therefore does not settle the ranking after repair.

\paragraph{Strong and weak models fail differently.}
Of \terra's 681 first-answer failures on Extended1000, 66.5\% are derailments and 17.0\% malformed answers; of \sol's 332, 47.0\% come from interactions between cars: collisions, wrong delivery order, or a car entering the wrong engine.
Hard200 shows the same split across providers (\cref{fig:curves}c): car interactions cause 59\% of \sol's failed first answers, 13--29\% of those of \deepseek, \minimax, and \haiku, and almost none of \terra's and \qwenbig's, which are mostly derailments, malformed answers, or rule-breaking edits.
Weaker models often fail to connect a car to its destination at all; stronger models fail on how routes interact (\cref{sec:q3}).

\paragraph{Format errors do not decide the ranking.}
Malformed answers make up a third of \terra's first answers on Hard200, and rule-breaking edits (building on occupied cells, adding disallowed pieces, or exceeding the budget) 46\% of \qwenbig's.
Scoring only first answers that are well formed and follow the rules raises \terra's \Sk{1} from 0.5\% to 0.8\% on Hard200 and from 31.9\% to 36.7\% on Extended1000, and leaves the order of all nine models unchanged (\suppref{sup:results}{B}).

\begin{figure}[t]
\centering
\includegraphics[width=\linewidth]{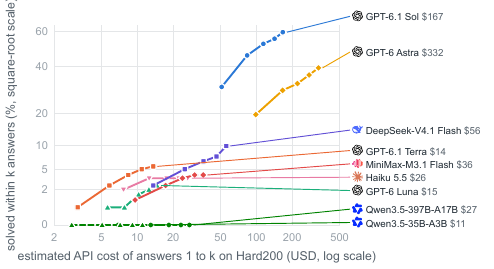}
\caption{\textbf{Estimated cost and success on Hard200.} Each line has five points: the estimated API cost of answers 1 to $k$ over all 200 puzzles, and \Sk{k}, for $k=1,\dots,5$. Costs apply each provider's official list price to the recorded tokens (off-peak price for \deepseek, prices in yuan converted at 7.0 per US dollar, no batch discount); they are estimates, not recorded spending. Labels give the cost at $k=5$.}
\label{fig:cost}
\end{figure}

\paragraph{Spending more does not buy planning.}
In \cref{fig:cost}, \sol reaches 59.5\% for about \$167, while \astra spends about \$332 for 39.5\%.
\deepseek reaches 10.0\% for about \$56, while \qwenbig spends about \$27 and solves nothing.
Within a model, each extra answer adds roughly the same cost but less success, so the curves flatten.
Long reasoning alone does not make up for a weak plan.

\paragraph{Mechanisms matter less than run length and crowding.}
Raw rates differ by mechanism (\suppref{sup:puzzles}{C}): every model does worse with passenger stations or post offices than with plain rails or tunnels.
But such puzzles also tend to be larger and to run longer.
Once a logistic regression on Extended1000 also includes board area, number of cars, track budget, run length, and search nodes, most mechanism effects for the four models shrink and lose significance: only stations and tunnels for \terra and self-switching points for \haiku remain harmful, placeable signals become helpful, and longer runs predict failure for every model except \sol.
A mechanism's cost thus lies mostly in the longer, more crowded runs it creates (\cref{sec:q3}).

%% file: sec/5_analysis.tex
\section{Diagnosing Planning and Repair}
\label{sec:analysis}

We now use the fixed panel to look inside \terra's planning and repair, asking in turn whether it can read the board, knows and can learn the rules, can predict how a given layout runs, what makes a layout hard to build, and whether feedback helps it repair (\cref{tab:panel}).

\begin{table}[t]
\caption{\textbf{Controlled conditions} (\terra, 180 puzzles, three runs pooled; \%). $\Delta$ compares \Sk{1} with the default image input for the input rows, \Sk{5} with the matched condition that states the rules (40.7) for the rules row, and \Sk{5} with three visual frames for the feedback rows. All feedback rows share the same first answers.}
\label{tab:panel}
\centering
\tablestyle\setlength{\tabcolsep}{3.4pt}
\begin{tabular}{@{}l ccc@{}}
\toprule
Condition & \Sk{1} & \Sk{5} & $\Delta$ [95\% CI] \\
\midrule
\multicolumn{4}{@{}l}{\emph{What the model sees}} \\
\quad Image (default) & 22.8 & 39.6 & -- \\
\quad Exact text board, text feedback & 24.3 & 41.1 & $+1.5$ [$-1.3$, $4.3$] \\
\quad Exact text board, visual feedback & 24.3 & 42.6 & -- \\
\quad Image + exact text layout & 25.0 & -- & $+2.2$ [$-1.1$, $5.6$] \\
\quad Image at $0.75\times$ size & 21.5 & -- & $-1.3$ [$-4.3$, $1.7$] \\
\quad Image at $1.5\times$ size & 24.1 & -- & $+1.3$ [$-1.7$, $4.3$] \\
\multicolumn{4}{@{}l}{\emph{What the model is told about the rules}} \\
\quad Mechanism rules withheld & 16.9 & 30.9 & $-9.8$ [$-13.3$, $-6.5$] \\
\multicolumn{4}{@{}l}{\emph{What the model gets after the same failure}} \\
\quad Fresh retry, no history & 22.8 & 36.5 & $-3.1$ [$-6.5$, $0.2$] \\
\quad Binary: only \texttt{failed} & 22.8 & 36.1 & $-3.5$ [$-6.1$, $-0.9$] \\
\quad Error type and cells & 22.8 & 35.4 & $-4.3$ [$-6.7$, $-2.0$] \\
\quad Text states, 3 & 22.8 & 37.8 & $-1.9$ [$-4.3$, $0.4$] \\
\quad Visual frames, 1 & 22.8 & 38.9 & $-0.7$ [$-2.6$, $1.1$] \\
\quad Visual frames, 3 (base) & 22.8 & 39.6 & -- \\
\quad Visual frames, 5 & 22.8 & 39.8 & $+0.2$ [$-1.7$, $2.2$] \\
\quad Visual frames, all & 22.8 & 39.1 & $-0.6$ [$-3.0$, $1.9$] \\
\quad 3 frames, 4 steps apart & 22.8 & 40.6 & $+0.9$ [$-0.9$, $2.8$] \\
\bottomrule
\end{tabular}
\end{table}

\subsection{Reading the Board}
\label{sec:q1}

A planner first needs to know the current state. Is reading it the main difficulty?
When asked only to copy the board, \terra reproduces every cell, mechanism, and car exactly for 97.2\% of panel puzzles (95\% CI 94.8--99.1).
Replacing the image with an exact text version of the board raises \Sk{1} from 22.8\% to 24.3\%, a change not distinguishable from zero.
With the same visual feedback, it raises \Sk{5} from 39.6\% to 42.6\%, a gain of 3.0 points (95\% CI 0.2--5.9).
Adding the exact text next to the image, or changing the image size, moves \Sk{1} by at most 2.2 points.
Exact knowledge of the board thus helps, but modestly: even with it, \terra solves fewer than half of the panel within five answers.

\subsection{Knowing and Learning the Rules}
\label{sec:q4}

A planner also needs to know how the world changes, from written rules or from examples on other boards.
In planning, hiding how the mechanisms behave while keeping the goal, the legend, and the basic movement rules lowers \Sk{1} from 22.8\% to 16.9\% and the matched \Sk{5} from 40.7\% to 30.9\% (\cref{tab:panel}).
The rules thus give the model knowledge it uses, not just names for what it sees.

To test whether the model can learn a rule from examples, we hide one mechanism's rule in the prediction task on 169 eligible puzzles and instead show two example runs on structurally different boards (\cref{fig:simulation}b).
With image input, examples of the same mechanism raise the share of fully correct predictions from 39.6\% to 64.1\%, close to the 69.0\% reached with the written rule.
Examples of a different mechanism give 38.7\%, no better than nothing.
With text input, the written rule still works better than matched examples (90.7\% versus 66.7\%).
So examples can teach how a mechanism behaves on a new board: written rules help planning, and examples help prediction.

\subsection{Predicting versus Planning}
\label{sec:q2}

\begin{figure}[t]
\centering
\includegraphics[width=\linewidth]{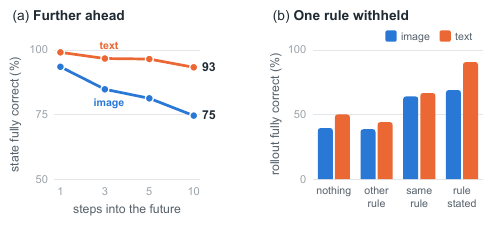}
\caption{\textbf{Predicting how a layout runs.} (a) Share of predicted board states that are fully correct, by how many steps ahead the model predicts, for reference layouts. (b) With one mechanism's rule withheld, share of predictions with all four requested states correct, when the prompt gives nothing, an example run of another mechanism, an example run of the same mechanism on a different board, or the written rule.}
\label{fig:simulation}
\end{figure}

Knowing the rules helps only if the model can work out their consequences.
We give the model a complete layout, either the reference solution or a failing layout, and ask for the positions and headings of all cars and the states of gates and switches after 1, 3, 5, and 10 steps.
For reference layouts, \terra gets all four states right in 91.3\% of cases with text input and 70.2\% with image input.
The gap grows the further ahead it predicts (\cref{fig:simulation}a): per-state accuracy is 99.1\% versus 93.5\% at step 1 and 93.3\% versus 74.6\% at step 10.
The model reads a still image well, but tracks a changing board from pixels less reliably.

A paired test on 22 puzzles changes one cell more than two cells from a car's start, either altering that car's later path (relevant) or not (control).
With text input, predictions are correct in 92.4\% of control cases but in 78.8\% after relevant changes, so the model often misses effects that travel across the board.
Still, predicting a given layout is far easier than building one: both scores are far above the 22.8\% first-attempt planning success.
The gap lies in using prediction to choose a plan that works.

\subsection{Combining Spatial and Timing Constraints}
\label{sec:q3}

\begin{figure}[t]
\centering
\includegraphics[width=\linewidth]{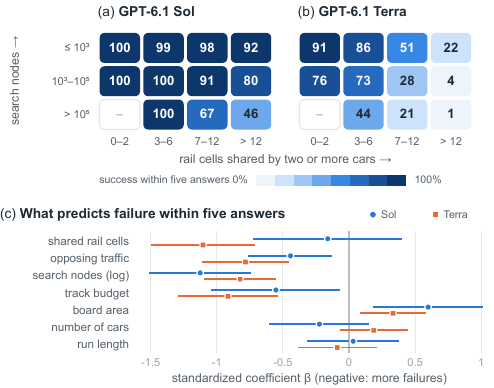}
\caption{\textbf{Shared routes predict failure beyond search effort.} (a, b) Success within five answers (\%) on Extended1000, grouped by solver search nodes (rows) and by the number of ordinary rail cells used by two or more required cars in the reference solution (columns). A dash marks a group with fewer than five puzzles. (c) Standardized coefficients of a logistic regression of the same outcome on puzzle properties, with 95\% intervals; negative values mean more failures.}
\label{fig:coupling}
\end{figure}

Building a layout means combining many constraints, each easy to understand.
We look at one concrete source of difficulty: how much the cars' routes interact, alongside solver search effort, track budget, board size, number of cars, and run length.
Among puzzles of medium search effort, \terra's \Sk{5} falls from 76\% when cars share at most two rail cells to 4\% when they share more than twelve, and \sol's from 100\% to 80\% (\cref{fig:coupling}).
The drop appears at every level of search effort.

A logistic regression on the 1,000 Extended1000 puzzles predicts \Sk{5} from standardized puzzle properties, including log search nodes and two measures of how the reference routes interact (\cref{fig:coupling}c).
For \terra, shared cells ($\beta=-1.11\pm0.20$) and opposing traffic ($\beta=-0.78\pm0.17$) predict failure about as strongly as search effort ($\beta=-0.82\pm0.14$); run length is not significant once the others are included.
For \sol, opposing traffic remains significant ($\beta=-0.44\pm0.16$) alongside search effort ($\beta=-1.13\pm0.20$).
Route interaction thus makes planning harder beyond what a solver must search.

\subsection{Repairing with Feedback}
\label{sec:q5}

\begin{figure}[t]
\centering
\includegraphics[width=\linewidth]{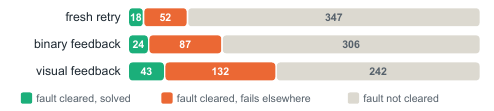}
\caption{\textbf{What the first revision does.} Outcome of the first revision of the same 417 failed answers. In the legend, \emph{fault cleared} means that the fault shown in the feedback no longer occurs; \emph{fault not cleared} also includes faults that are bypassed or cannot be compared.}
\label{fig:feedback}
\end{figure}

Does feedback raise success, beat simply trying again, and fix the fault it shows?
All feedback conditions start from the same first answers (\cref{tab:panel}).
Three visual frames give 39.6\% success, against 36.1\% for binary feedback and 36.5\% for five fresh attempts.
More or fewer frames, or frames spaced differently, stay within one point of this.
The error type and failing cells alone give 35.4\%.
The form of feedback matters, but more detail does not translate into larger gains.

To separate feedback from extra computation, we compare success at equal tokens per puzzle (input plus output, which includes reasoning).
Fresh sampling may use up to 64 attempts; feedback conditions allow five answers.
At 40k tokens, fresh retries solve 40.6\%, visual feedback 39.1\%, and binary feedback 36.1\%.
The paired difference between fresh retries and visual feedback is 1.5 points (95\% CI $[-1.9, 4.8]$), so neither is reliably better at this budget.

Using feedback well does not guarantee a working plan.
Visual feedback clears the fault in 175 of 417 first revisions, versus 70 for fresh retries, and its edits land closer to the fault: 47\% of changed cells are on or next to a fault cell, versus 35\%.
Yet 132 of these 175 revisions fail somewhere else, leaving 43 that solve the puzzle (\cref{fig:feedback}).
\Cref{fig:teaser} shows a typical case of one local fix after another.

%% file: sec/6_conclusion.tex
\section{Discussion and Conclusion}
\label{sec:conclusion}

\paragraph{Knowing the parts is not planning.}
The diagnostics separate what a model knows from what it does with that knowledge.
\terra reads the board almost perfectly, learns a mechanism from a few examples, and predicts how a given layout runs, yet it rarely builds a layout that works.
The missing step is to use these pieces together: to imagine several cars moving at once on a layout that does not exist yet, and to keep every route consistent with the others.
This is why failures follow how much the routes interact, not only how much a solver must search, and why stronger models, which reliably connect cars to their destinations, fail mainly through collisions and delivery order.

\paragraph{Repair treats the symptom.}
Feedback draws edits to the place where the run failed, and the edit usually removes that failure.
But the reported fault is often a symptom of routes that do not fit together, so a local fix moves the conflict elsewhere (\cref{fig:teaser}), and at an equal token budget visual feedback is no better than starting over.
Repair should be judged both by whether the reported fault is fixed and by whether the revised plan works, and compared with fresh retries at equal cost.

\paragraph{Implications.}
Spending more does not close the gap.
Models that plan in dynamic visual settings need to check a candidate plan against the joint motion it will produce before committing to it, and feedback that explains why a plan fails, not only where; \bench measures both with an exact simulator, and Hard200 leaves ample room above the strongest model.

\paragraph{Limitations.}
The controlled diagnostics use one model; stronger models fail differently on the leaderboard, and their diagnostics may differ.
All puzzles originate from a single domain, with rules defined by our simulator that vary in certain aspects from the original game.
Difficulty labels and interaction measures come from one solver and one reference solution per puzzle.
Closed models may change behind the same name; we record exact identifiers, dates, and settings.
We did not measure human performance.
\bench separates obtaining and predicting task information from using it to build and repair plans that work; we release the puzzles, simulator, solver, and evaluation code\ifarxiv{} as open source on GitHub (\codeurl)\fi{} to support progress on both abilities.

%% file: sec/supp.tex
\ifarxiv
The appendices give the model settings and protocol details (\cref{sup:settings}), additional results (\cref{sup:results}), the puzzle sets and the mechanism analysis (\cref{sup:puzzles}), the rules of the simulator (\cref{sup:rules}), the diagnostic studies (\cref{sup:diagnostics}), the prompts and feedback messages (\cref{sup:prompts}), related evaluation settings (\cref{sup:related}), and the data and records behind every number (\cref{sup:data}).
\else
This supplementary material gives the model settings and protocol details (\cref{sup:settings}), additional results (\cref{sup:results}), the puzzle sets and the mechanism analysis (\cref{sup:puzzles}), the rules of the simulator (\cref{sup:rules}), the diagnostic studies (\cref{sup:diagnostics}), the prompts and feedback messages (\cref{sup:prompts}), related evaluation settings (\cref{sup:related}), and the data and records behind every number (\cref{sup:data}). Section, table and figure numbers without the prefix S refer to the main paper.
\fi

\section{Model settings and protocol details}
\label{sup:settings}

\begin{table*}[t]
\caption{\textbf{Model configurations.} ``Both'' means Hard200 and Extended1000. ChatGPT: requests made with a ChatGPT account instead of an API key; the service sets the output limit itself, so no maximum is stated for it.}
\label{tab:model-config}
\centering
\tablestyle
\begin{tabular}{@{}lllll@{}}
\toprule
Model & Puzzle sets & Service & Reasoning & Output limit per generation \\
\midrule
\sol & Both & Azure OpenAI; ChatGPT & high & 128,000 tokens; service-managed \\
\astra & Both & ChatGPT & high & service-managed \\
\terra & Both & ChatGPT & high & service-managed \\
\luna & Both & Azure OpenAI; ChatGPT & high & 128,000 tokens; service-managed \\
\haiku & Both & Anthropic API & high, adaptive & 128,000 tokens; continuations \\
\deepseek & Hard200 & DeepSeek API & thinking, maximum effort & model maximum \\
\minimax & Hard200 & MiniMax API (via Model Studio) & high & 524,288 tokens \\
\qwenbig & Hard200 & Alibaba Cloud Model Studio & default thinking & 65,536 tokens \\
\qwen & Hard200 & Alibaba Cloud Model Studio & default thinking & 65,536 tokens \\
\bottomrule
\end{tabular}
\end{table*}

\paragraph{Models and requests.}
The API identifiers are \texttt{gpt-6.1-sol}, \texttt{gpt-6-astra}, \texttt{gpt-6.1-terra}, \texttt{gpt-6-luna}, \texttt{claude-haiku-5-5}, \texttt{deepseek-flash}, \texttt{MiniMax/MiniMax-M3.1-Flash}, \texttt{qwen3.5-397b-a17b}, and \texttt{qwen3.5-35b-a3b} (\cref{tab:model-config}).
\minimax is called through the OpenAI-compatible endpoint of Alibaba Cloud Model Studio, which routes the request to the MiniMax API.
The leaderboard runs took place between 5 and 17 November 2026.
Temperature, nucleus sampling and sampling seeds are left at each service's default.
Every leaderboard run uses the prompt with the board image and the mechanism rules and returns additions to the given layout (\cref{sup:prompts}).
Each puzzle allows up to five graded answers and stops after a success.
After a failure, the model receives the failure description and at most three simulated state images, four steps apart; a simulation stops after at most 10,000 steps.
No model has tools, code execution, or access to the reference solution.
Between graded answers, the conversation keeps the final answer, not hidden reasoning.
An engineering failure, interrupted stream, or service error is not a graded answer.

\paragraph{Output limits.}
\haiku uses native adaptive thinking at high effort.
A generation that stops at the output limit does not count as an answer: it is either continued in plain text, if the model supports this (here \haiku, at most twice, keeping that answer's reasoning blocks and signatures, with no simulator feedback until the completed answer is graded), or requested again.
On Hard200, 78 graded \haiku answers used one continuation, and on Extended1000, 419 did; none needed a second.
Nine \luna generations on Hard200 and 201 \haiku generations on Extended1000, 200 of them with no final text, were requested again, as were empty replies from any model; no other model stopped at its limit.
Every graded answer in every run ended with a normal stop reported by the service, including the services that set the limit themselves, so no graded answer reached its output limit.
Five graded answers therefore need not mean five generations.

\paragraph{Cost estimates.}
The estimated cost of a run sums, over every request with reported usage, the uncached input tokens times the input price, the cached input tokens times the cached-input price, and the output tokens (including reasoning) times the output price, using the list prices of \cref{tab:prices}.
It is an estimate from list prices, not recorded spending.

\paragraph{Statistics.}
Confidence intervals for success rates come from resampling puzzles with replacement (percentile bootstrap, 10,000 resamples).
For the panel, the three runs of a puzzle are resampled together, so the intervals reflect the 180 puzzles rather than 540 independent observations.
For the 40k-token comparison in \mainref{sec:q5}{Sec.~5.5}, we average the paired fresh-minus-visual success differences over each puzzle's three runs and compute a 95\% percentile interval from 100,000 bootstrap resamples of the 180 puzzles (NumPy \texttt{default\_rng}, seed 20261117).
The logistic regressions are fitted by Newton's method on all 1,000 Extended1000 puzzles.
Every predictor is standardized to mean zero and unit variance over these puzzles, standard errors come from the inverse of the fitted information matrix, and the pseudo-$R^2$ is McFadden's: one minus the ratio of the fitted and intercept-only log-likelihoods.
A coefficient is marked when it is at least twice its standard error ($|z|\geq2$).

\section{Additional results}
\label{sup:results}

\begin{table*}[t]
\caption{\textbf{Answer format, sensitivity, and sampling uncertainty} (\%). Malformed: first answers that cannot be parsed or have the wrong structure. Rule-breaking: first answers that build on occupied cells, add pieces that may not be added, or exceed the track budget. No graded answer in any run was cut off at the output limit. Valid only: \Sk{1} among puzzles whose first answer is neither malformed nor rule-breaking (number of such puzzles in parentheses). Brackets: 95\% percentile bootstrap intervals over puzzles (10,000 resamples).}
\label{tab:robust}
\centering
\tablestyle
\begin{tabular}{@{}llrrcccc@{}}
\toprule
Set & Model & Malformed & Rule-breaking & \Sk{1} & \Sk{1}, valid only & \Sk{5} & Rep. \\
\midrule
Hard200 & \sol & 0.0 & 2.5 & 30.5 [24.5, 37.0] & 31.3 (195) & 59.5 [52.5, 66.0] & 41.7 [33.6, 50.0] \\
Hard200 & \astra & 0.0 & 9.0 & 19.5 [14.0, 25.0] & 21.4 (182) & 39.5 [33.0, 46.5] & 24.8 [18.4, 31.7] \\
Hard200 & \deepseek & 0.5 & 17.5 & 2.5 [0.5, 5.0] & 3.0 (164) & 10.0 [6.0, 14.5] & 7.7 [4.1, 11.7] \\
Hard200 & \terra & 33.0 & 2.0 & 0.5 [0.0, 1.5] & 0.8 (130) & 5.5 [2.5, 9.0] & 5.0 [2.0, 8.1] \\
Hard200 & \minimax & 2.0 & 26.0 & 1.0 [0.0, 2.5] & 1.4 (144) & 4.0 [1.5, 7.0] & 3.0 [1.0, 5.6] \\
Hard200 & \haiku & 9.0 & 8.5 & 2.0 [0.5, 4.0] & 2.4 (165) & 3.5 [1.0, 6.0] & 1.5 [0.0, 3.6] \\
Hard200 & \luna & 3.0 & 14.0 & 0.5 [0.0, 1.5] & 0.6 (166) & 2.5 [0.5, 5.0] & 2.0 [0.5, 4.0] \\
Hard200 & \qwenbig & 0.5 & 45.5 & 0.0 [0.0, 0.0] & 0.0 (108) & 0.0 [0.0, 0.0] & 0.0 [0.0, 0.0] \\
Hard200 & \qwen & 13.0 & 42.5 & 0.0 [0.0, 0.0] & 0.0 (89) & 0.0 [0.0, 0.0] & 0.0 [0.0, 0.0] \\
\midrule
Ext.1000 & \sol & 0.0 & 4.3 & 66.8 [63.8, 69.8] & 69.8 (957) & 90.9 [89.0, 92.6] & 72.6 [67.7, 77.3] \\
Ext.1000 & \astra & 0.0 & 5.5 & 59.6 [56.6, 62.7] & 63.1 (945) & 83.5 [81.2, 85.8] & 59.2 [54.3, 63.9] \\
Ext.1000 & \terra & 11.6 & 1.5 & 31.9 [29.0, 34.8] & 36.7 (869) & 50.0 [46.9, 53.1] & 26.6 [23.3, 30.0] \\
Ext.1000 & \haiku & 4.9 & 5.3 & 26.6 [23.9, 29.4] & 29.6 (898) & 50.9 [47.8, 53.9] & 33.1 [29.7, 36.5] \\
Ext.1000 & \luna & 6.1 & 15.0 & 7.5 [5.9, 9.2] & 9.5 (789) & 16.1 [13.8, 18.4] & 9.3 [7.4, 11.2] \\
\bottomrule
\end{tabular}
\end{table*}

\paragraph{Answer format and uncertainty.}
\Cref{tab:robust} separates format and rule errors from planning failures.
Scoring only valid first answers raises \Sk{1} by at most 4.8 points and leaves the order of the models unchanged on both sets.
Each leaderboard model is run once; the intervals in \cref{tab:robust} reflect which puzzles are sampled, not repeated runs.
On the 180-puzzle panel, three independent \terra runs of the base condition give \Sk{1} of 22.2, 21.7, and 24.4\% and \Sk{5} of 38.9, 41.1, and 38.9\% (\cref{tab:panel-reps}).
The intervals of the seven models at or below 10\% on Hard200 overlap, so their order should not be over-read.

\begin{table}[t]
\caption{\textbf{Exact cumulative solved counts} $N_k$, the number of puzzles solved within $k$ answers. Hard200 has 200 puzzles and Extended1000 has 1,000. The repair rate is $(N_5-N_1)/(N-N_1)$.}
\label{tab:all-rounds}
\centering
\tablestyle
\begin{tabular}{@{}llrrrrr@{}}
\toprule
Set & Model & $N_1$ & $N_2$ & $N_3$ & $N_4$ & $N_5$ \\
\midrule
Hard200 & \sol & 61 & 92 & 105 & 111 & 119 \\
Hard200 & \astra & 39 & 58 & 64 & 72 & 79 \\
Hard200 & \deepseek & 5 & 10 & 13 & 15 & 20 \\
Hard200 & \terra & 1 & 5 & 8 & 10 & 11 \\
Hard200 & \minimax & 2 & 5 & 7 & 8 & 8 \\
Hard200 & \haiku & 4 & 7 & 7 & 7 & 7 \\
Hard200 & \luna & 1 & 1 & 3 & 4 & 5 \\
Hard200 & \qwenbig & 0 & 0 & 0 & 0 & 0 \\
Hard200 & \qwen & 0 & 0 & 0 & 0 & 0 \\
\midrule
Ext.1000 & \sol & 668 & 802 & 864 & 898 & 909 \\
Ext.1000 & \astra & 596 & 735 & 787 & 809 & 835 \\
Ext.1000 & \haiku & 266 & 411 & 471 & 493 & 509 \\
Ext.1000 & \terra & 319 & 418 & 451 & 479 & 500 \\
Ext.1000 & \luna & 75 & 116 & 140 & 156 & 161 \\
\bottomrule
\end{tabular}
\end{table}

\begin{table}[t]
\caption{\textbf{Extended1000 \Sk{5} by solver difficulty} (\%).}
\label{tab:extension-difficulty}
\centering
\tablestyle
\begin{tabular}{@{}lccc@{}}
\toprule
Model & Easy (598) & Medium (300) & Hard (102) \\
\midrule
\sol & 98.2 & 88.3 & 55.9 \\
\astra & 94.5 & 74.3 & 46.1 \\
\haiku & 68.4 & 28.0 & 15.7 \\
\terra & 67.7 & 28.0 & 10.8 \\
\luna & 25.4 & 3.0 & 0.0 \\
\bottomrule
\end{tabular}
\end{table}

\paragraph{First-answer failures.}
\Cref{tab:fail-types} counts what goes wrong in each failed first answer; \mainref{fig:curves}{Fig.~4c} shows the same counts grouped into four categories, in which car interactions are collisions, wrong delivery order, and a car entering the wrong engine, and a decoy entering an engine counts among the other errors.
The simulator stops a run at its first failure, so each failed answer is counted once, under the first failure that occurs.
The failure types are defined in \cref{tab:failure-defs}.

\begin{table*}[t]
\caption{\textbf{What goes wrong in failed first answers} (counts). Collision includes collisions inside a tunnel pair. Order: wrong delivery order. Engine: a car enters an engine of the other type. Decoy: a decoy enters an engine. Stop: a car reaches its engine before serving every stop assigned to it. Rules: an edit that breaks the placement rules. Budget: more new rails than the budget. $n$: failed first answers.}
\label{tab:fail-types}
\centering
\tablestyle\setlength{\tabcolsep}{3.0pt}
\begin{tabular}{@{}ll rrrrrrrrrr r@{}}
\toprule
Set & Model & Derailment & Collision & Order & Engine & Decoy & Stop & Loop & Malformed & Rules & Budget & $n$ \\
\midrule
Hard200 & \sol & 50 & 64 & 15 & 3 & 0 & 1 & 1 & 0 & 0 & 5 & 139 \\
Hard200 & \astra & 54 & 71 & 15 & 3 & 0 & 0 & 0 & 0 & 4 & 14 & 161 \\
Hard200 & \deepseek & 119 & 30 & 6 & 2 & 0 & 1 & 1 & 1 & 32 & 3 & 195 \\
Hard200 & \terra & 124 & 4 & 1 & 0 & 0 & 0 & 0 & 66 & 1 & 3 & 199 \\
Hard200 & \minimax & 116 & 21 & 5 & 0 & 0 & 0 & 0 & 4 & 40 & 12 & 198 \\
Hard200 & \haiku & 103 & 28 & 26 & 2 & 1 & 1 & 0 & 18 & 4 & 13 & 196 \\
Hard200 & \luna & 128 & 29 & 6 & 0 & 0 & 0 & 2 & 6 & 25 & 3 & 199 \\
Hard200 & \qwenbig & 102 & 5 & 1 & 0 & 0 & 0 & 0 & 1 & 75 & 16 & 200 \\
Hard200 & \qwen & 89 & 0 & 0 & 0 & 0 & 0 & 0 & 26 & 79 & 6 & 200 \\
\midrule
Ext.1000 & \sol & 119 & 133 & 18 & 5 & 1 & 1 & 12 & 0 & 6 & 37 & 332 \\
Ext.1000 & \astra & 164 & 143 & 23 & 4 & 3 & 0 & 12 & 0 & 2 & 53 & 404 \\
Ext.1000 & \terra & 453 & 58 & 24 & 3 & 4 & 0 & 8 & 116 & 6 & 9 & 681 \\
Ext.1000 & \haiku & 213 & 206 & 174 & 7 & 1 & 2 & 29 & 49 & 19 & 34 & 734 \\
Ext.1000 & \luna & 579 & 67 & 35 & 5 & 0 & 1 & 27 & 61 & 137 & 13 & 925 \\
\bottomrule
\end{tabular}
\end{table*}

\paragraph{Hard200 by puzzle property.}
\Cref{tab:hard-props} splits Hard200 into thirds by six properties for the two models that solve a substantial share of it.
Success falls with the track budget, the run length, and both measures of route interaction (defined in \cref{sup:puzzles}); it changes less with the number of cars and is lowest on the largest boards.
On the main set, too, puzzles whose required cars share more rail cells are solved less often: \sol solves 76.6\% of the third with the fewest shared cells and 47.8\% of the third with the most, and \astra 57.1\% and 20.9\%.

\begin{table}[t]
\caption{\textbf{Hard200 \Sk{5} by thirds of a puzzle property} (\%). Each property splits the 200 puzzles into three groups of about equal size, from low (L) to high (H); ties keep some groups unequal.}
\label{tab:hard-props}
\centering
\tablestyle\setlength{\tabcolsep}{3.4pt}
\begin{tabular}{@{}l ccc ccc@{}}
\toprule
& \multicolumn{3}{c}{\sol} & \multicolumn{3}{c}{\astra} \\
\cmidrule(lr){2-4}\cmidrule(l){5-7}
Property & L & M & H & L & M & H \\
\midrule
Track budget & 77.8 & 52.5 & 41.7 & 60.5 & 27.1 & 23.3 \\
Run length & 72.1 & 53.8 & 52.2 & 58.8 & 30.8 & 28.4 \\
Shared rail cells & 76.6 & 50.0 & 47.8 & 57.1 & 37.5 & 20.9 \\
Opposing shared edges & 72.9 & 57.8 & 47.0 & 50.0 & 40.6 & 27.3 \\
Cars & 62.6 & 62.2 & 50.0 & 41.1 & 44.4 & 31.2 \\
Board cells & 62.0 & 65.6 & 50.8 & 46.5 & 45.3 & 26.2 \\
\bottomrule
\end{tabular}
\end{table}

\begin{figure*}[t]
\centering
\includegraphics[width=\textwidth]{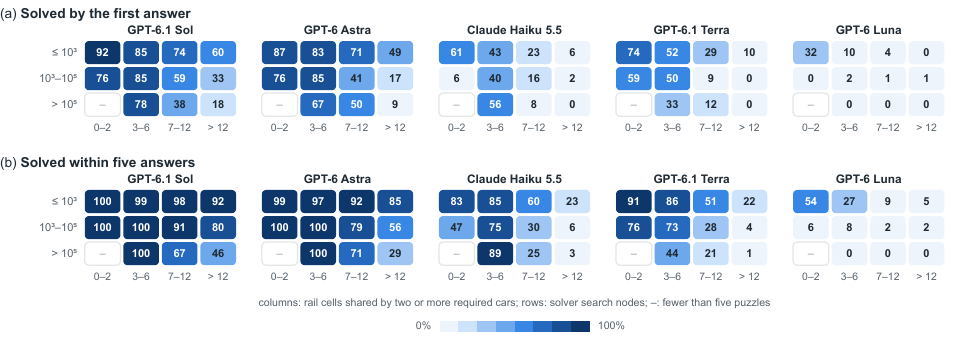}
\caption{\textbf{Success by search effort and route sharing for every model with a full Extended1000 run} (\%). Rows group puzzles by the solver's search nodes, columns by the number of ordinary rail cells used by two or more required cars in the reference solution. (a) First answer; (b) within five answers.}
\label{fig:supp-heat}
\end{figure*}

\paragraph{Route interaction across models.}
\mainref{fig:coupling}{Fig.~7} shows \sol and \terra at five answers.
\Cref{fig:supp-heat} adds the first answer and the other three models with a full Extended1000 run, and \cref{tab:coupling-all} gives the corresponding regressions, with the same predictors as in \mainref{sec:q3}{Sec.~5.4}.
For each of the four models other than \luna, puzzles in which the required cars share more than twelve rail cells are solved least often at every level of search effort, for the first answer and within five answers.
With the other properties held fixed, shared rail cells predict failure for \astra, \haiku and \terra at both points, and opposing traffic does so for \sol, \astra and \terra at both points and for \haiku within five answers.
\luna solves too few puzzles for precise estimates.
Run length predicts failure for \astra, \haiku and \terra in the mechanism model of \cref{tab:mech-adjusted}, which has no route-interaction measures; once shared rail cells and opposing traffic enter (\cref{tab:coupling-all}), it remains a significant predictor only for \haiku.

\begin{table*}[t]
\caption{\textbf{Puzzle properties that predict success on Extended1000.} Standardized logistic-regression coefficients ($\pm$ standard error) for solving a puzzle with the first answer and within five answers; negative values mean more failures. Search nodes enter as $\log(1+n)$; the other predictors are counts. Bold: $|z|\geq2$.}
\label{tab:coupling-all}
\centering
\tablestyle
\begin{tabular}{@{}lccccc@{}}
\toprule
& \sol & \astra & \haiku & \terra & \luna \\
\midrule
\multicolumn{6}{@{}l}{\emph{Solved by the first answer}} \\
\quad Shared rail cells & $-0.23{\pm}0.17$ & $\mathbf{-0.48{\pm}0.17}$ & $\mathbf{-0.83{\pm}0.24}$ & $\mathbf{-0.77{\pm}0.22}$ & $-0.91{\pm}0.58$ \\
\quad Opposing shared edges & $\mathbf{-0.31{\pm}0.12}$ & $\mathbf{-0.45{\pm}0.13}$ & $+0.10{\pm}0.19$ & $\mathbf{-0.63{\pm}0.21}$ & $-0.33{\pm}0.55$ \\
\quad Search nodes (log) & $\mathbf{-0.67{\pm}0.12}$ & $\mathbf{-0.65{\pm}0.12}$ & $\mathbf{-0.54{\pm}0.15}$ & $\mathbf{-0.75{\pm}0.15}$ & $\mathbf{-1.70{\pm}0.45}$ \\
\quad Track budget & $\mathbf{-0.60{\pm}0.16}$ & $\mathbf{-0.61{\pm}0.16}$ & $-0.34{\pm}0.19$ & $\mathbf{-0.86{\pm}0.20}$ & $-0.25{\pm}0.40$ \\
\quad Run length & $-0.10{\pm}0.12$ & $-0.19{\pm}0.13$ & $\mathbf{-1.30{\pm}0.29}$ & $-0.36{\pm}0.21$ & $-0.93{\pm}0.61$ \\
\quad Board cells & $\mathbf{+0.36{\pm}0.11}$ & $\mathbf{+0.34{\pm}0.11}$ & $\mathbf{+0.47{\pm}0.13}$ & $\mathbf{+0.36{\pm}0.13}$ & $\mathbf{-0.64{\pm}0.28}$ \\
\quad Required cars & $+0.18{\pm}0.12$ & $\mathbf{+0.25{\pm}0.12}$ & $+0.01{\pm}0.15$ & $+0.20{\pm}0.14$ & $+0.21{\pm}0.30$ \\
\quad Pseudo-$R^2$ & 0.21 & 0.27 & 0.26 & 0.35 & 0.41 \\
\multicolumn{6}{@{}l}{\emph{Solved within five answers}} \\
\quad Shared rail cells & $-0.16{\pm}0.29$ & $\mathbf{-0.51{\pm}0.22}$ & $\mathbf{-0.46{\pm}0.20}$ & $\mathbf{-1.11{\pm}0.20}$ & $-0.42{\pm}0.33$ \\
\quad Opposing shared edges & $\mathbf{-0.44{\pm}0.16}$ & $\mathbf{-0.34{\pm}0.13}$ & $\mathbf{-0.47{\pm}0.16}$ & $\mathbf{-0.78{\pm}0.17}$ & $-0.37{\pm}0.31$ \\
\quad Search nodes (log) & $\mathbf{-1.13{\pm}0.20}$ & $\mathbf{-0.87{\pm}0.15}$ & $\mathbf{-0.86{\pm}0.13}$ & $\mathbf{-0.82{\pm}0.14}$ & $\mathbf{-1.44{\pm}0.28}$ \\
\quad Track budget & $\mathbf{-0.55{\pm}0.25}$ & $-0.32{\pm}0.19$ & $-0.19{\pm}0.19$ & $\mathbf{-0.92{\pm}0.19}$ & $\mathbf{-1.76{\pm}0.31}$ \\
\quad Run length & $+0.03{\pm}0.18$ & $-0.06{\pm}0.14$ & $\mathbf{-1.47{\pm}0.25}$ & $-0.09{\pm}0.15$ & $-0.02{\pm}0.27$ \\
\quad Board cells & $\mathbf{+0.60{\pm}0.21}$ & $+0.30{\pm}0.15$ & $\mathbf{+0.39{\pm}0.13}$ & $\mathbf{+0.33{\pm}0.13}$ & $-0.21{\pm}0.18$ \\
\quad Required cars & $-0.22{\pm}0.19$ & $-0.03{\pm}0.15$ & $-0.08{\pm}0.14$ & $+0.19{\pm}0.13$ & $\mathbf{+0.52{\pm}0.19}$ \\
\quad Pseudo-$R^2$ & 0.34 & 0.30 & 0.38 & 0.45 & 0.44 \\
\bottomrule
\end{tabular}
\end{table*}

\begin{table}[t]
\caption{\textbf{Recorded tokens and estimated cost} of the full runs (five answers). Tokens in millions, from all records with returned usage; output includes reasoning. Cost in USD at the list prices of \cref{tab:prices}; an estimate, not recorded spending.}
\label{tab:token-use}
\centering
\tablestyle
\begin{tabular}{@{}llrrr@{}}
\toprule
Set & Model & Input & Output & Cost \\
\midrule
Hard200 & \sol & 5.38 & 16.08 & 167 \\
Hard200 & \astra & 6.46 & 5.38 & 332 \\
Hard200 & \deepseek & 5.03 & 97.35 & 56 \\
Hard200 & \terra & 7.06 & 1.49 & 14 \\
Hard200 & \minimax & 7.50 & 60.87 & 36 \\
Hard200 & \haiku & 20.11 & 48.89 & 26 \\
Hard200 & \luna & 7.70 & 28.41 & 15 \\
Hard200 & \qwenbig & 5.24 & 25.73 & 27 \\
Hard200 & \qwen & 4.45 & 23.59 & 11 \\
\midrule
Ext.1000 & \sol & 7.12 & 14.98 & 163 \\
Ext.1000 & \astra & 9.30 & 9.99 & 591 \\
Ext.1000 & \haiku & 79.60 & 223.67 & 120 \\
Ext.1000 & \terra & 16.51 & 5.27 & 41 \\
Ext.1000 & \luna & 21.52 & 15.91 & 10 \\
\bottomrule
\end{tabular}
\end{table}

\begin{table}[t]
\caption{\textbf{List prices used for cost estimates}, per million tokens, standard tier, no batch discount, checked 2026-11-17. Prices for \deepseek (off-peak), \minimax, and the Qwen models are the official prices in yuan (CNY) and are converted at 7.0 CNY per USD; the others are official prices in USD. Cached input applies to tokens reported as cache hits.}
\label{tab:prices}
\centering
\tablestyle
\begin{tabular}{@{}lcccc@{}}
\toprule
Model & Currency & Input & Cached input & Output \\
\midrule
\sol & USD & 2.00 & 0.10 & 10.00 \\
\astra & USD & 10.00 & 1.00 & 50.00 \\
\terra & USD & 1.00 & 0.05 & 5.00 \\
\luna & USD & 0.10 & 0.01 & 0.50 \\
\haiku & USD & 0.10 & 0.01 & 0.50 \\
\deepseek & CNY & 1.00 & 0.02 & 4.00 \\
\minimax & CNY & 1.00 & 0.10 & 4.00 \\
\qwenbig & CNY & 1.20 & -- & 7.20 \\
\qwen & CNY & 0.40 & -- & 3.20 \\
\bottomrule
\end{tabular}
\end{table}

\section{Puzzle sets and mechanisms}
\label{sup:puzzles}

\paragraph{What a puzzle contains.}
A puzzle is a rectangular board with fixed rails, mechanisms and blocked cells; the cars, each with a start cell, a heading, a type and a number; one or two engines; and two budgets, the most new rail cells and the most new signals the answer may add.
Rocks and station buildings both block travel and rail placement, so all comparisons below treat them as the same blocked cell.
All 1,200 Extended1000 and Hard200 puzzles meet the following conditions: every car starts on a straight rail cell with no mechanism; gates lie on ordinary straight or curved rails, never on three-way rails; and passenger stations and post offices lie on straight rails.
\Cref{fig:examples} shows five puzzles with their reference solutions.

\paragraph{How the puzzles were made.}
\ifarxiv
Coding agents (OpenAI Codex and Anthropic Claude Code) proposed puzzles and reference solutions under constraints set by the authors.
\else
As stated in the LLM usage statement, coding agents proposed puzzles and reference solutions under constraints set by the authors.
\fi
A proposal was kept only if the simulator accepted its reference solution, the reference layout met the conditions above and below, the solver could set its budgets, and the authors accepted the images of the puzzle and of its solved layout.
No answer from an evaluated model was used to select puzzles.

\paragraph{Reference solutions and budgets.}
Every designed puzzle has a reference solution that the simulator accepts.
In each reference layout, every rail end connects to a matching rail, except the unused rear end of a starting car, and the layout uses ordinary two-way rails where an extra branch of a three-way rail would serve no purpose.
The track budget equals the number of new rail cells in the reference solution, which is the fewest the solver found.
For Extended1000 this minimum is proved for 987 puzzles, 970 by exhausting the search and 17 because the solution matches a lower bound (a count of new rails that every solution must contain); the other 13 budgets come from the best solutions found within 600 seconds.
For Hard200 it is proved for 136 puzzles.
The authors reviewed the image of every puzzle and of its solved layout, removed puzzles that differ from another only in empty space or decoration, and kept no puzzle with the same gameplay in two sets.

\paragraph{Search solver and difficulty labels.}
The solver is an exact search over layouts that runs the simulator as it goes; it adapts the search of RailboundSolver~\cite{railboundsolver} to our rules.
Whenever a car is about to enter an empty cell, it branches over the pieces that could be placed there; each such choice is one \emph{search node}.
It drops a branch as soon as a move fails, the budget runs out, a reachability check shows that some required car can no longer reach its stops and engine, or a lower bound shows that the branch cannot beat a solution already found.
Every solution it returns is replayed by the benchmark's own validator.
Difficulty is the number of search nodes visited before the first solution when the search starts without the reference solution, with a 600-second allowance and no node cap.
This search solves 992 of the 1,000 Extended1000 puzzles (median 470 nodes) and 229 of the 233 original levels; the remaining puzzles keep their reference solutions.
Every one of the 233 original levels has a solution within its original track allowance that our simulator accepts.
For Hard200 the difficulty check stops at 100,001 nodes: all 200 puzzles reach that limit, and the 16 that were also searched without a node limit needed between $1.8\times10^5$ and $8.5\times10^6$ nodes.
These labels measure the search effort of this solver, not difficulty for people or models.

\paragraph{The three sets.}
\Cref{tab:corpus-compare} compares Hard200, Extended1000 and the 233 original levels.
Extended1000 has nearly the same statistics as the original levels: the difficulty mix by solver search effort is almost the same (59.8/30.0/10.2\% against 58.4/31.3/10.3\% easy/medium/hard), the median board, budget and run length are close, and most mechanisms appear in a similar share of puzzles; buttons and button-switched points are less common in Extended1000, and plain-rail puzzles more common.
Hard200 has larger budgets, longer runs, more cars and more mechanisms per puzzle, and in almost all of its puzzles the required cars share rail cells (99.5\%) and travel some shared rail in opposite directions (90.0\%).

\begin{table}[t]
\caption{\textbf{The three puzzle sets.} Sizes as median (minimum--maximum); mechanisms as the share of puzzles that contain them (\%). Shared rail cells: ordinary rail cells used by two or more required cars in the reference solution; opposing shared edges: connections between two neighboring cells that different required cars travel in opposite directions.}
\label{tab:corpus-compare}
\centering
\tablestyle\setlength{\tabcolsep}{3.2pt}
\begin{tabular}{@{}lccc@{}}
\toprule
& Hard200 & Extended1000 & Original \\
\midrule
Puzzles & 200 & 1,000 & 233 \\
Easy / medium / hard & 0 / 0 / 200 & 598 / 300 / 102 & 136 / 73 / 24 \\
Board cells & 50.5 (24--182) & 30 (6--108) & 32 (5--99) \\
Cars, incl.\ decoys & 4 (1--10) & 3 (1--6) & 2 (1--5) \\
Track budget & 19 (9--62) & 11 (1--43) & 11 (0--48) \\
Run length (steps) & 43 (12--439) & 17 (3--153) & 18 (4--188) \\
Shared rail cells & 22.5 (0--77) & 8 (0--43) & 8 (0--46) \\
Opposing shared edges & 5.5 (0--37) & 1 (0--16) & 1 (0--18) \\
Mechanisms per puzzle & 4 (0--10) & 2 (0--11) & 2 (0--11) \\
Any blocked cell (\%) & 6.5 & 18.6 & 36.9 \\
Any shared rail cell (\%) & 99.5 & 89.6 & 88.4 \\
Any opposing traffic (\%) & 90.0 & 57.8 & 53.6 \\
\midrule
Plain rails only & 0.5 & 15.3 & 9.9 \\
Fixed three-way rails & 21.5 & 13.5 & 15.9 \\
Buttons & 59.5 & 41.5 & 52.4 \\
Gates & 44.5 & 35.2 & 37.3 \\
Button-switched points & 29.5 & 15.9 & 26.6 \\
Self-switching points & 82.5 & 30.4 & 28.8 \\
Tunnels & 23.5 & 19.9 & 24.0 \\
Passenger stations & 47.5 & 24.9 & 22.3 \\
Post offices & 12.0 & 3.3 & 4.3 \\
Fixed signals & 2.0 & 1.9 & 0.4 \\
Placeable signals & 38.0 & 12.0 & 15.9 \\
Two engines & 43.0 & 18.7 & 16.7 \\
Decoy cars & 6.5 & 8.5 & 13.3 \\
\bottomrule
\end{tabular}
\end{table}

\paragraph{No repeated puzzles.}
We compared every pair among the 1,433 puzzles of the three sets directly.
Each puzzle is written as a grid of cell contents: the fixed rail piece and its orientation, the mechanism with the car number of a station or post office, blocked cells, and the starting car with its type, heading and number; numbers that only link cells, such as button groups and tunnel pairs, are renumbered in reading order, and empty border rows and columns are removed.
Two puzzles have the same gameplay when these grids are equal under one of the eight rotations and reflections and both budgets are equal.
No Extended1000 or Hard200 puzzle has the same gameplay as an original level or as any other puzzle in the three sets.

\paragraph{Random layouts.}
For every puzzle we drew 200 random layouts: a number $n$ chosen uniformly between one and the track budget, $n$ empty cells chosen uniformly, and a uniformly random rail piece (straight, curved or three-way) on each; no signals were placed, and each layout was run for at most 2,000 steps.
Of the 200,000 Extended1000 layouts, 18 (0.009\%) solve their puzzle, all on four easy puzzles with a budget of one rail; none of the 40,000 Hard200 layouts does, and 8 of the 46,600 layouts for the original levels do, on two easy puzzles.
Almost all random layouts derail (99.3\% for Extended1000).

\paragraph{Controlled panel.}
The 180 panel puzzles were chosen from Extended1000 by their properties alone, without model results.
Within each difficulty level, a deterministic greedy selection first fills coverage minimums, at least $\min(4, \text{available})$ puzzles for every mechanism and at least one puzzle for every number of cars (1 to 4, and 5 or more) and every number of mechanisms (0 to 3, and 4 or more), and then maximizes a diversity score; one-for-one exchanges that keep every minimum and raise the score are applied until none remains.
The panel holds 60 easy, 59 medium and 61 hard puzzles.
It contains 97 puzzles with buttons, 93 with self-switching points, 80 with gates, 68 with two engines, 62 with passenger stations, 45 with tunnels, 42 with button-switched points, 41 with placeable signals, 34 with fixed three-way rails, 29 with post offices, 29 with decoys, 17 with fixed signals, and 4 with plain rails only.

\begin{figure*}[t]
\centering
\setlength{\tabcolsep}{2pt}
\begin{tabular}{@{}ccccc@{}}
\includegraphics[height=2.38cm]{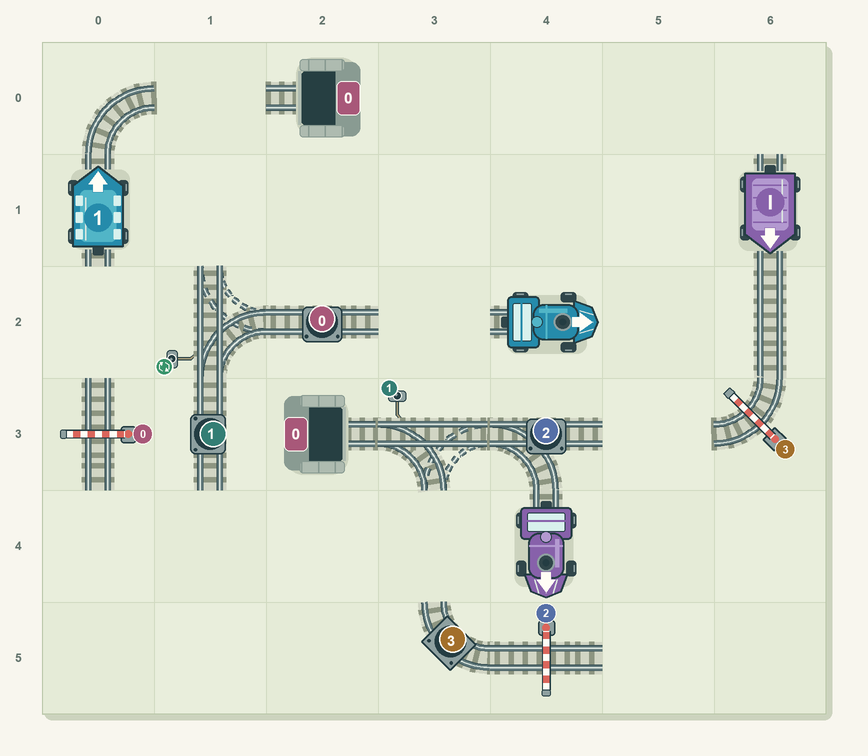} &
\includegraphics[height=2.38cm]{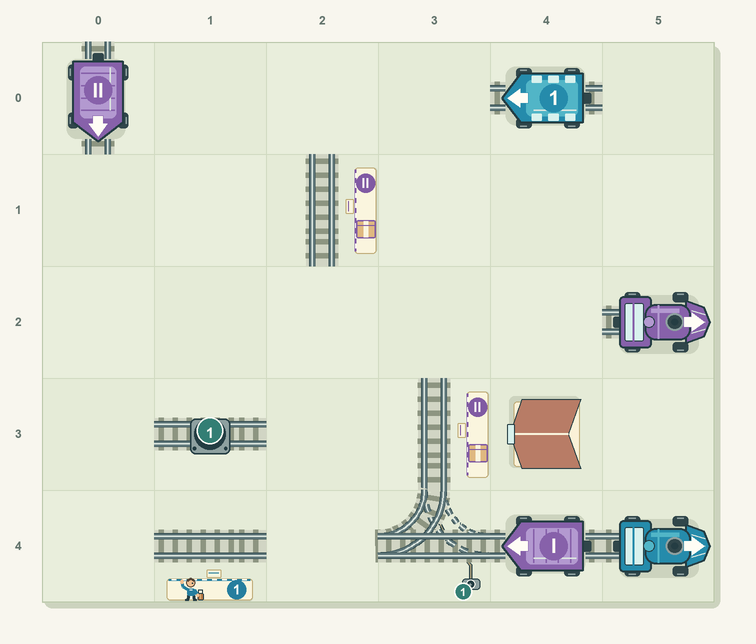} &
\includegraphics[height=2.38cm]{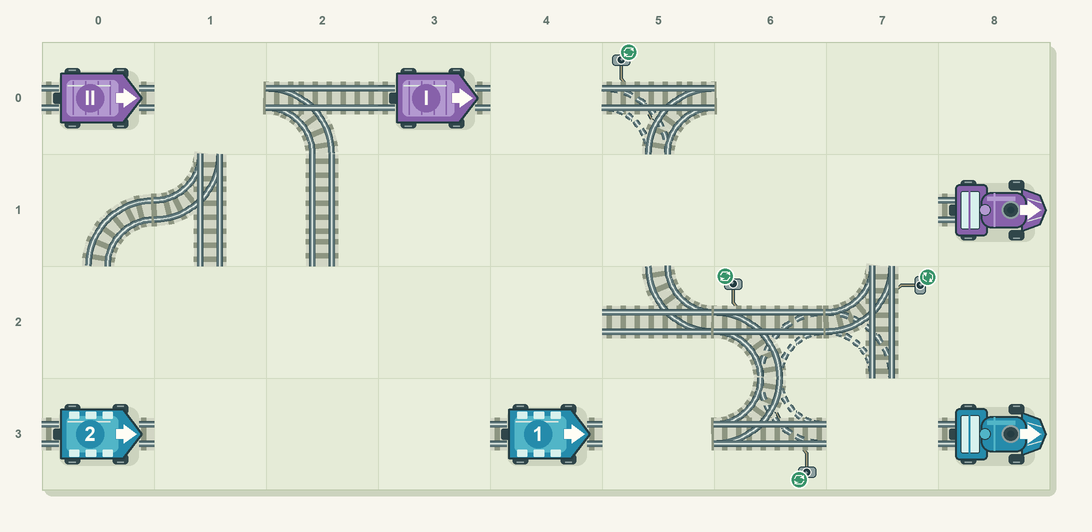} &
\includegraphics[height=2.38cm]{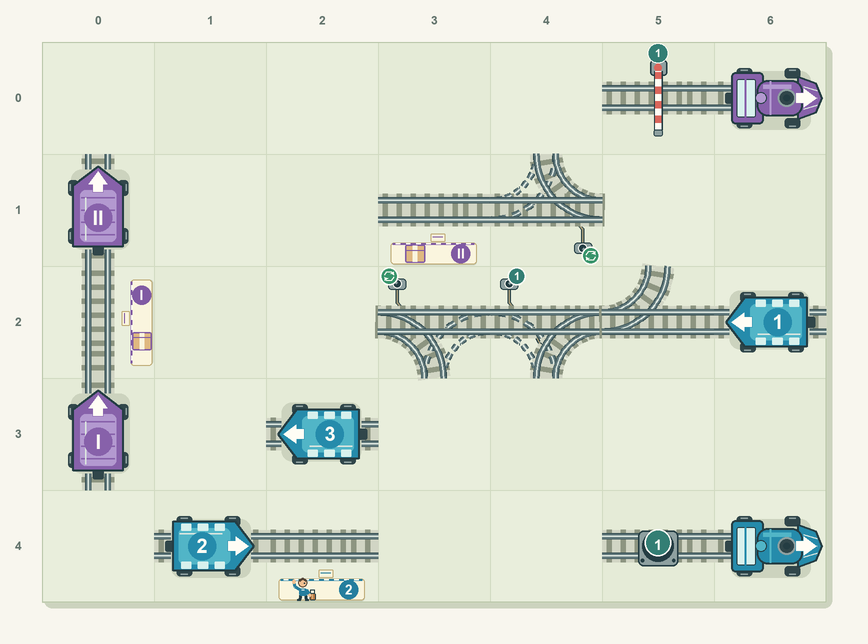} &
\includegraphics[height=2.38cm]{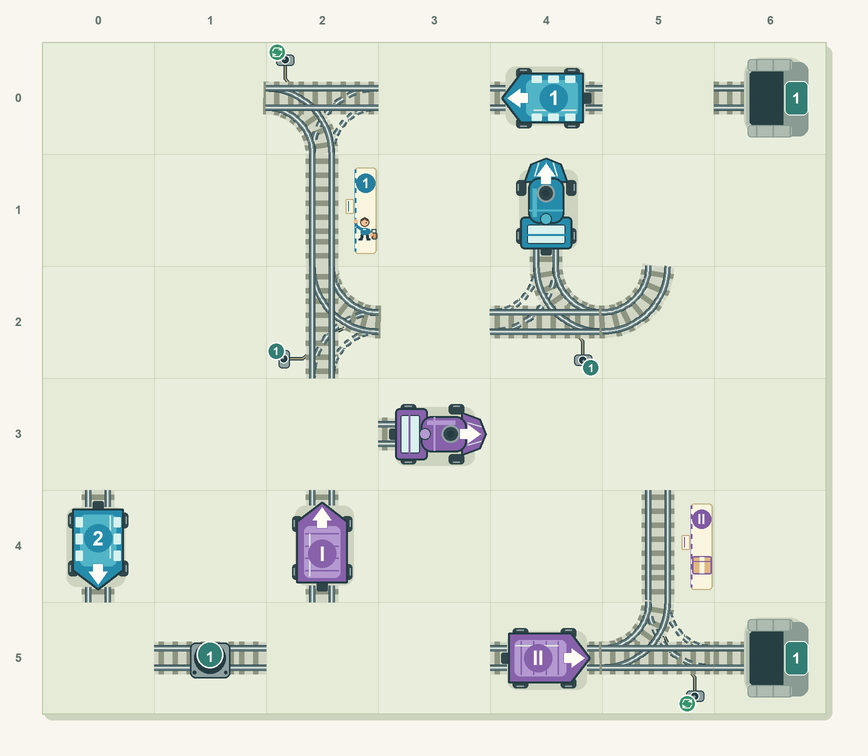} \\
\includegraphics[height=2.38cm]{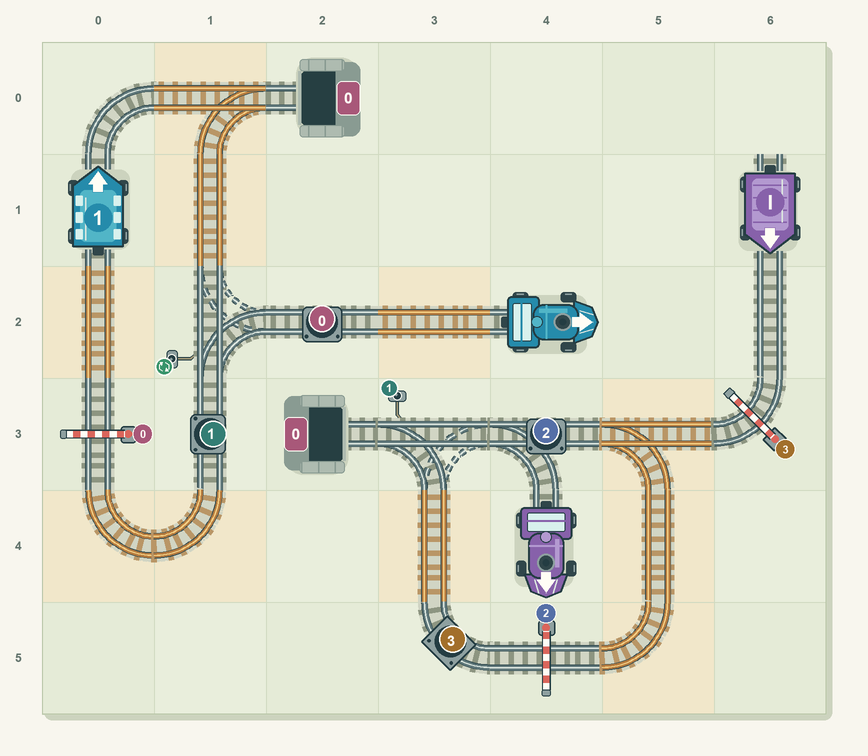} &
\includegraphics[height=2.38cm]{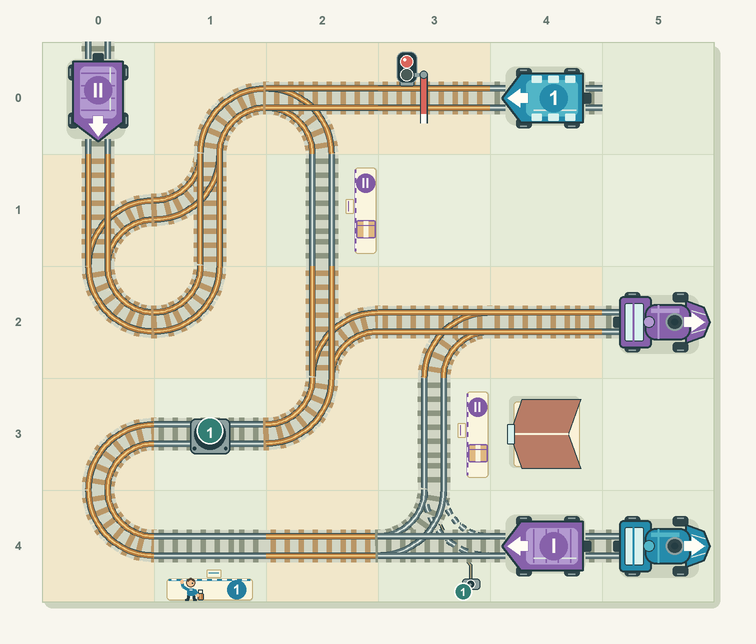} &
\includegraphics[height=2.38cm]{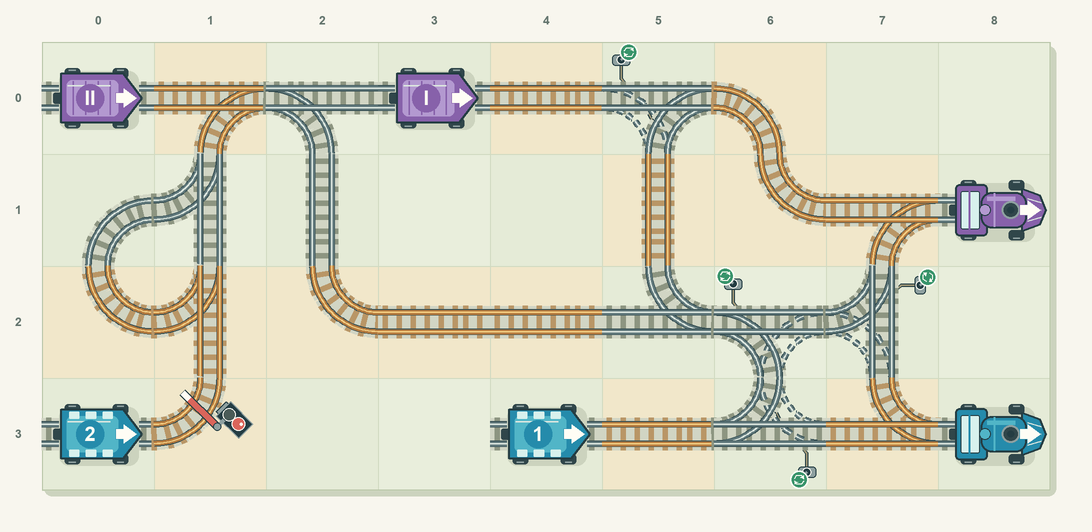} &
\includegraphics[height=2.38cm]{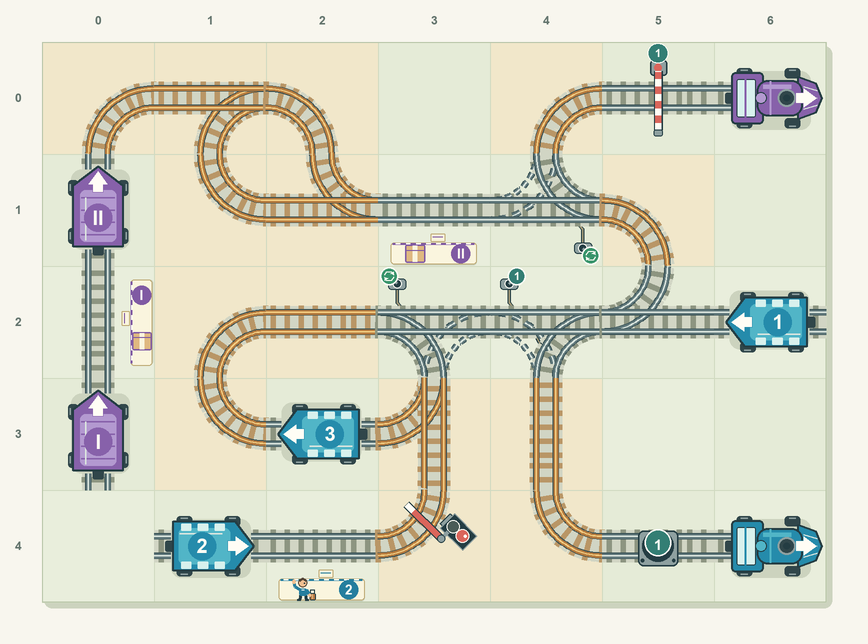} &
\includegraphics[height=2.38cm]{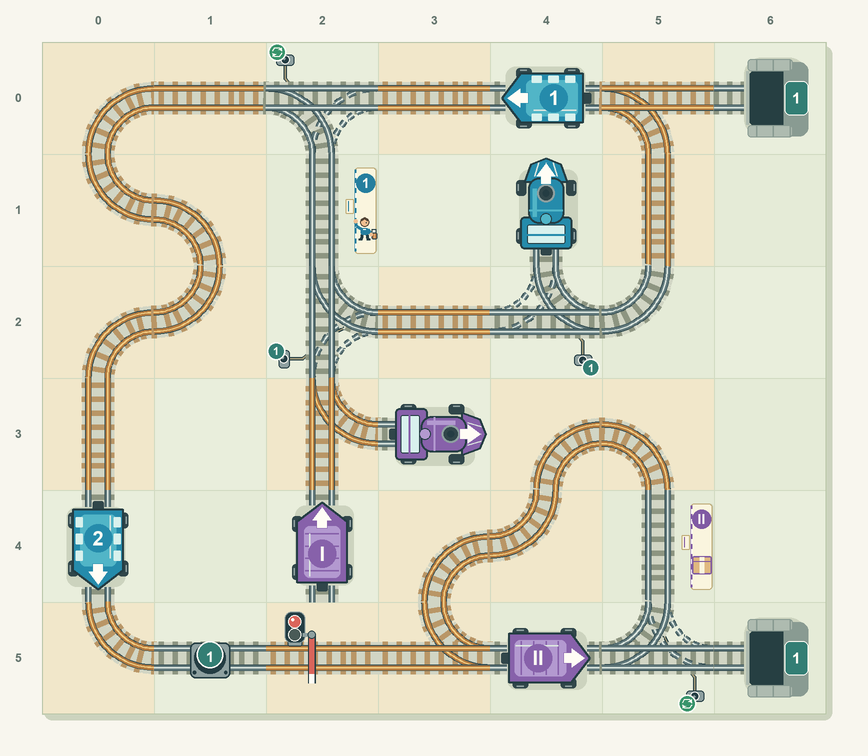} \\[2pt]
\scriptsize (a) Extended1000, easy & \scriptsize (b) Extended1000, medium & \scriptsize (c) Extended1000, hard & \scriptsize (d) Hard200 & \scriptsize (e) Hard200 \\
\end{tabular}
\caption{\textbf{Example puzzles} (top) and their reference solutions (bottom), as rendered for the models. (a) Tunnels, buttons, gates and two engines; budget 10. (b) Passenger stations, post offices, a placeable signal and two engines; budget 14. (c) Self-switching points, a placeable signal and two engines; budget 14. (d) Nine mechanisms and five cars; budget 14. (e) Four cars whose reference run lasts 148 steps; budget 19.}
\label{fig:examples}
\end{figure*}

\paragraph{Mechanism groups.}
Mechanism groups are computed from the board, the mechanisms, the cars and the budgets; a puzzle may belong to several groups.
Fixed three-way rails are three-way rails on the board that never switch, that is, without a button-switched or self-switching mechanism.
The plain-rail group excludes fixed three-way rails, buttons, gates, switching points, tunnels, stations and post offices, fixed or placeable signals, multiple engine types, and decoys.
The ``Buttons'' row contains all 415 puzzles with buttons, whereas 352 puzzles contain gates.
\Cref{tab:mechanisms} gives the raw \Sk{5} rates by mechanism, and \cref{tab:mech-adjusted} the effects after controlling for puzzle size and search effort.

\begin{table}[t]
\caption{\textbf{Extended1000 \Sk{5} by mechanism} (\%). A puzzle counts in every row whose element it contains, so groups overlap. Plain rails: no listed mechanism and no fixed signal. Two engines: separate passenger and freight engines. Raw rates, not adjusted for puzzle size; \cref{tab:mech-adjusted} gives the adjusted effects.}
\label{tab:mechanisms}
\centering
\tablestyle\setlength{\tabcolsep}{3.2pt}
\begin{tabular}{@{}lr ccccc@{}}
\toprule
Mechanism & $n$ & Sol & Astra & Haiku & Terra & Luna \\
\midrule
Plain rails only & 153 & 98.0 & 91.5 & 71.9 & 70.6 & 39.2 \\
Tunnels & 199 & 96.5 & 95.0 & 69.3 & 67.8 & 18.6 \\
Decoy cars & 85 & 100.0 & 91.8 & 44.7 & 65.9 & 5.9 \\
Fixed three-way rails & 135 & 91.9 & 79.3 & 48.1 & 53.3 & 26.7 \\
Placeable signals & 120 & 91.7 & 88.3 & 59.2 & 56.7 & 12.5 \\
Buttons & 415 & 89.4 & 80.2 & 44.3 & 49.9 & 10.8 \\
Two engines & 187 & 86.6 & 77.5 & 48.7 & 41.2 & 13.9 \\
Button-switched points & 159 & 89.9 & 75.5 & 35.8 & 39.6 & 10.7 \\
Self-switching points & 304 & 87.2 & 76.3 & 32.9 & 40.5 & 11.2 \\
Passenger stations & 249 & 83.9 & 73.9 & 34.1 & 27.7 & 8.0 \\
Post offices & 33 & 75.8 & 57.6 & 30.3 & 21.2 & 12.1 \\
\bottomrule
\end{tabular}
\end{table}

\begin{table*}[t]
\caption{\textbf{Adjusted mechanism effects on Extended1000 \Sk{5}.} Coefficients ($\pm$ standard error) of a logistic regression that also includes the logarithms of board area, number of cars, track budget, run length, and search nodes (standardized). Bold: $|z|\geq2$. \sol solves all 85 decoy puzzles, so this coefficient has no finite estimate for it.}
\label{tab:mech-adjusted}
\centering
\tablestyle\setlength{\tabcolsep}{3pt}
\begin{tabular}{@{}lcccc@{}}
\toprule
& Sol & Astra & Haiku & Terra \\
\midrule
Run length (log) & $-0.39{\pm}.25$ & $\mathbf{-0.54{\pm}.18}$ & $\mathbf{-1.56{\pm}.19}$ & $\mathbf{-1.23{\pm}.18}$ \\
Search nodes (log) & $\mathbf{-1.15{\pm}.25}$ & $\mathbf{-0.99{\pm}.19}$ & $\mathbf{-1.03{\pm}.16}$ & $\mathbf{-0.95{\pm}.17}$ \\
\midrule
Fixed three-way rails & $+0.07{\pm}.46$ & $-0.63{\pm}.35$ & $+0.11{\pm}.32$ & $-0.34{\pm}.34$ \\
Buttons & $-1.19{\pm}.79$ & $-0.54{\pm}.56$ & $-0.90{\pm}.51$ & $-0.04{\pm}.51$ \\
Gates & $+0.06{\pm}.71$ & $-0.10{\pm}.49$ & $+0.40{\pm}.45$ & $-0.08{\pm}.45$ \\
Button-switched points & $+0.59{\pm}.55$ & $-0.29{\pm}.39$ & $+0.01{\pm}.35$ & $-0.50{\pm}.36$ \\
Self-switching points & $-0.03{\pm}.37$ & $-0.13{\pm}.28$ & $\mathbf{-0.52{\pm}.23}$ & $-0.07{\pm}.24$ \\
Tunnels & $-0.21{\pm}.49$ & $+0.71{\pm}.42$ & $-0.13{\pm}.25$ & $\mathbf{-0.53{\pm}.26}$ \\
Passenger stations & $-0.42{\pm}.34$ & $-0.17{\pm}.26$ & $-0.23{\pm}.24$ & $\mathbf{-0.60{\pm}.24}$ \\
Post offices & $+0.52{\pm}.69$ & $-0.30{\pm}.57$ & $-0.19{\pm}.62$ & $-0.47{\pm}.64$ \\
Placeable signals & $+0.51{\pm}.50$ & $\mathbf{+1.23{\pm}.43}$ & $\mathbf{+0.96{\pm}.33}$ & $\mathbf{+0.95{\pm}.33}$ \\
Two engines & $-0.36{\pm}.39$ & $-0.03{\pm}.31$ & $+0.28{\pm}.27$ & $-0.30{\pm}.27$ \\
Decoy cars & -- & $+0.30{\pm}.56$ & $-0.15{\pm}.39$ & $+0.29{\pm}.42$ \\
\bottomrule
\end{tabular}
\end{table*}

\paragraph{Route-interaction measures.}
Both measures are read from the run of the reference solution in the simulator.
A rail cell is \emph{shared} when two or more required cars (decoys excluded) pass through it; only ordinary rail cells count, not engines or tunnel portals.
An edge, the connection between two neighboring cells, has \emph{opposing traffic} when different required cars travel along it in opposite directions.
Because a puzzle may have other solutions, both measures describe the reference solution, not every way to solve the puzzle.

\section{Rules of the simulator}
\label{sup:rules}

\paragraph{One step.}
All cars move at once, one cell per step, along the rail of the cell they enter; a car's heading is the direction of its next move.
Within a step the simulator works in this order.
(1)~A gate that was waiting to close because a car stood in it closes once the cell is empty.
(2)~Cars that entered a cell in the previous step act on it: a button press toggles every gate and button-switched point with the same number, a self-switching point flips its curve after the entering car has taken its heading, and a car that passes the three-way rail next to a closed signal, heading away from the signal, opens that signal for good.
(3)~Every moving car proposes its next move.
A car stopped at its own station waits two extra steps, and a car in front of a closed gate or closed signal waits until it opens.
A car entering a tunnel portal from its open side leaves the paired portal in the same step.
A car reaching an engine is delivered if it has the engine's type, comes from the engine's open side, is the lowest undelivered number of its type, and has served every stop assigned to it.
(4)~Conflicting proposals are failures: two cars in one cell, two cars swapping cells head-on, a car entering the cell of a waiting car, or two cars passing through the same tunnel pair in opposite directions.
A car may follow another into the cell it is leaving in the same step.
A decoy that derails stays as a stationary wreck that blocks its cell; two decoys may collide without failing the run.
(5)~A delivered car leaves the board at the end of its move, and the other cars continue.
The run succeeds when every required car has been delivered (after two more steps if decoys are still moving), fails when the complete state repeats an earlier one, a loop check that stops runs which would otherwise repeat forever, and stops after 10,000 steps, a limit that no answer in any run reached.

\paragraph{Differences from the original game.}
Our simulator follows the rules of Railbound~\cite{railbound} with two changes: engines may stand inside the board, and a delivered car leaves at the end of its move without pausing the other cars.
All results follow our simulator's rules.

\begin{table}[t]
\caption{\textbf{Failure types} returned by the simulator and the answer checker. The names are those shown to the model in feedback.}
\label{tab:failure-defs}
\centering
\tablestyle
\begin{tabular}{@{}lp{0.6\linewidth}@{}}
\toprule
Name & Meaning \\
\midrule
\texttt{derailment} & A car runs off the board, onto an empty or incompatible cell, into the closed side of a tunnel, or into an engine from the wrong side. \\
\texttt{collision} & Two cars in one cell, swapping cells, or one entering a waiting car's cell. \\
\texttt{tunnel\_collision} & Two cars use one tunnel pair in opposite directions. \\
\texttt{wrong\_order} & A car reaches its engine before a lower-numbered car of its type. \\
\texttt{wrong\_locomotive} & A car enters an engine of the other type. \\
\texttt{decoy\_locomotive} & A decoy enters an engine. \\
\texttt{unvisited\_station} & A car reaches its engine before serving all its stops. \\
\texttt{loop} & The run returns to an earlier state. \\
\texttt{invalid\_json} & The reply is not one JSON object. \\
\texttt{invalid\_schema} & The JSON has wrong keys, values or positions. \\
\texttt{illegal\_edit} & The answer builds on an occupied cell, adds a piece that may not be added, or places a signal where none is allowed. \\
\texttt{track\_budget} & More new rail cells than the budget. \\
\texttt{semaphore\_budget} & More new signals than the budget. \\
\bottomrule
\end{tabular}
\end{table}

\section{Diagnostic study details}
\label{sup:diagnostics}

All diagnostic studies use \terra on the 180-puzzle panel, three times per condition (\mainref{sec:puzzles}{Sec.~3.2}).

\paragraph{Every condition and repetition.}
\Cref{tab:panel-reps} lists \Sk{1} and \Sk{5} for each run of each condition; \mainref{tab:panel}{Tab.~3} pools the three runs.
Conditions with the same first input share their first answers within a run, so their \Sk{1} values coincide.
The base condition shows three frames, one step apart; the leaderboard shows three frames four steps apart.
Feedback frames end at the failure, so a failure in the first steps yields fewer frames, and an answer that cannot be run yields none.
Across the four frame counts and the five other combinations of frame count and spacing, \Sk{5} differs from the base condition by $-1.3$ to $+1.1$ points, and the 95\% paired bootstrap interval of every difference includes zero.

\begin{table}[t]
\caption{\textbf{Every panel condition, by run} (\%; runs 1 / 2 / 3). Copying the board is scored as the share of puzzles whose transcription is exact. Rules stated (matched): the counterpart of the withheld-rules condition, with rules stated and the same feedback text, which leaves out the sentences explaining the feedback overlays. Dashes: single-answer conditions.}
\label{tab:panel-reps}
\centering
\tablestyle\setlength{\tabcolsep}{3pt}
\begin{tabular}{@{}lcc@{}}
\toprule
Condition & \Sk{1} & \Sk{5} \\
\midrule
Copy the board (exact) & 97.2 / 97.2 / 97.2 & -- \\
Image (default) & 22.2 / 21.7 / 24.4 & 38.9 / 41.1 / 38.9 \\
Image + exact text layout & 23.3 / 29.4 / 22.2 & -- \\
Image at 0.75$\times$ size & 21.1 / 22.2 / 21.1 & -- \\
Image at 1.5$\times$ size & 23.9 / 23.9 / 24.4 & -- \\
Exact text board, text feedback & 22.2 / 25.6 / 25.0 & 41.7 / 43.3 / 38.3 \\
Exact text board, visual feedback & 22.2 / 25.6 / 25.0 & 43.3 / 44.4 / 40.0 \\
Rules stated (matched) & 22.2 / 21.7 / 24.4 & 39.4 / 42.8 / 40.0 \\
Mechanism rules withheld & 15.6 / 18.3 / 16.7 & 31.7 / 30.0 / 31.1 \\
Fresh retry, no history & 22.2 / 21.7 / 24.4 & 37.2 / 35.6 / 36.7 \\
Binary: only \texttt{failed} & 22.2 / 21.7 / 24.4 & 35.6 / 36.7 / 36.1 \\
Error type and cells & 22.2 / 21.7 / 24.4 & 35.0 / 35.0 / 36.1 \\
Text states, 3 & 22.2 / 21.7 / 24.4 & 35.0 / 40.0 / 38.3 \\
Visual frames, 1 & 22.2 / 21.7 / 24.4 & 36.7 / 42.2 / 37.8 \\
Visual frames, 3 (base) & 22.2 / 21.7 / 24.4 & 38.9 / 41.1 / 38.9 \\
Visual frames, 5 & 22.2 / 21.7 / 24.4 & 38.3 / 41.1 / 40.0 \\
Visual frames, 10 & 22.2 / 21.7 / 24.4 & 36.1 / 39.4 / 40.6 \\
Visual frames, all & 22.2 / 21.7 / 24.4 & 37.8 / 40.6 / 38.9 \\
3 frames, 2 steps apart & 22.2 / 21.7 / 24.4 & 37.8 / 38.3 / 38.9 \\
3 frames, 4 steps apart & 22.2 / 21.7 / 24.4 & 39.4 / 43.9 / 38.3 \\
5 frames, 2 steps apart & 22.2 / 21.7 / 24.4 & 41.7 / 40.0 / 40.6 \\
5 frames, 4 steps apart & 22.2 / 21.7 / 24.4 & 39.4 / 41.1 / 40.6 \\
\bottomrule
\end{tabular}
\end{table}

\paragraph{Reading the board.}
In the copying task the model sees the same image and description of the drawing as in planning and returns the complete board: every rail and mechanism code, every mechanism number, and every car with its position, heading, type and number.
A transcription counts as exact only if all of these match.
Exact copies are not enough to plan: in the 525 puzzle runs in which \terra copied the board exactly, its first layout for the same puzzle in the same run failed 406 times (77.3\%).
Planning answers also show few signs of misreading which cells are occupied: only 6 of \terra's 1,000 first answers on Extended1000 build on an occupied cell or add a piece that may not be added (\cref{tab:fail-types}).

\paragraph{Predicting a layout.}
The prediction task gives the model the puzzle and a complete layout and asks for the state after 1, 3, 5 and 10 steps: the position, heading and status of every car, and the contents of every cell that can change, such as switching points, gates, signals and served stops.
If the run ends earlier, the model also gives the step at which it ends and whether it succeeds or fails.
The layouts are the 180 reference solutions and 180 failing layouts: 150 failed answers that the model gave in the panel runs and 30 reference solutions with one new rail on a car's route removed.
Layouts whose run loops are not used.
A state counts as correct only if every requested value matches the simulator, and a query is correct if all four states are.
\Cref{tab:prediction} gives both scores.
In 486 of the 540 reference queries the run is still going at step 10, so all four states must be predicted; text input gets all four right in 90.3\% of these, and image input in 67.3\%.

\begin{table}[t]
\caption{\textbf{Predicting how a layout runs} (\terra, \%, three runs pooled; 540 queries per row). Per step: share of predicted states that are fully correct. All four: queries with every state correct.}
\label{tab:prediction}
\centering
\tablestyle\setlength{\tabcolsep}{3.4pt}
\begin{tabular}{@{}llccccc@{}}
\toprule
Input & Layout & Step 1 & Step 3 & Step 5 & Step 10 & All four \\
\midrule
Text & Reference & 99.1 & 96.7 & 96.5 & 93.3 & 91.3 \\
Text & Failing & 98.7 & 94.8 & 93.0 & 90.7 & 90.2 \\
Image & Reference & 93.5 & 84.8 & 81.3 & 74.6 & 70.2 \\
Image & Failing & 93.5 & 89.6 & 85.0 & 82.4 & 79.8 \\
\bottomrule
\end{tabular}
\end{table}

\paragraph{Changes far from a car.}
For the paired test, a puzzle is eligible when one cell more than two cells (in grid steps) from a car's start can be changed while the rails around that start stay the same; 22 panel puzzles qualify.
A relevant change alters that car's later path; the control change is of the same kind, at a different distant position, and does not.
Each query is scored as before, correct only if all four states are right. With text input, 92.4\% of queries are correct after the control change and 78.8\% after the relevant change; with image input, 53.0\% and 57.6\%, so image-based prediction is unreliable whether or not the change matters (66 queries per cell: 22 puzzles, three runs).

\paragraph{Rules from examples.}
For 169 eligible prediction queries, one mechanism present on the board loses its written rule: stations and post offices (44 queries), buttons with gates and button-switched points (34), self-switching points (17), signals (32), or tunnels (42).
The prompt then gives nothing, two example runs of another mechanism, two example runs of the hidden mechanism, or its written rule.
In every query, the hidden mechanism acts within the ten predicted steps.
Example runs come from Extended1000 puzzles outside the panel whose rail networks differ from the query board's, and are shown in the same state format as the query.
\Cref{tab:transfer} gives the share of queries with all four states correct.

\begin{table}[t]
\caption{\textbf{Prediction with one mechanism's rule withheld} (\terra, \%, 507 queries per cell: 169 queries, three runs).}
\label{tab:transfer}
\centering
\tablestyle
\begin{tabular}{@{}lcccc@{}}
\toprule
Input & Nothing & Other mechanism & Same mechanism & Written rule \\
\midrule
Image & 39.6 & 38.7 & 64.1 & 69.0 \\
Text & 50.1 & 44.2 & 66.7 & 90.7 \\
\bottomrule
\end{tabular}
\end{table}

\begin{figure}[t]
\centering
\includegraphics[width=\linewidth]{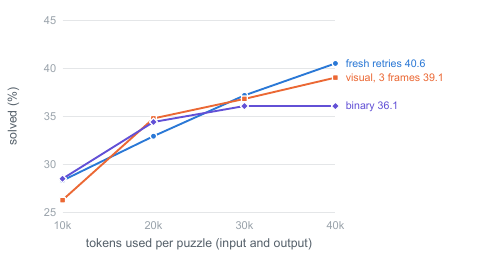}
\caption{\textbf{Success against tokens used} on the panel (\terra, 540 puzzle runs). Labels give the value at 40k tokens.}
\label{fig:budget}
\end{figure}

\paragraph{Feedback at an equal token budget.}
For each panel puzzle and run we follow three policies: fresh retries, which draw new independent first answers with no feedback, up to 64 per puzzle; visual feedback with three frames; and binary feedback, each with at most five answers.
A puzzle counts as solved at budget $B$ if a correct answer arrives before the tokens used on that puzzle, input plus output with reasoning, exceed $B$.
\Cref{fig:budget} shows the result from 10k to 40k tokens.
Feedback is ahead at 20k tokens, the policies cross between 20k and 30k, and binary feedback does not improve beyond 30k.

\paragraph{What the first revision changes.}
\Cref{tab:revisions} compares the first revision after each of the same 417 failed first answers under three policies.
A fault is \emph{cleared} when the revised layout runs the same car safely through the place or mechanism where it failed before; a revision may also route the car so that it never reaches that place, repeat the same failure there, or fail earlier, before the comparison can be made.
No comparison is possible when the failed answer reports no fault cell or one of the two layouts cannot be run.
Changed cells compare the revised layout with the failed one, and a changed cell is near the fault when it lies on or next to a reported fault cell.
\mainref{fig:teaser}{Fig.~1} reports the 43 of 417 visual-feedback revisions (10.3\%) that clear the fault and solve the puzzle; of the 13 other revisions that solve it, 11 route the car around the old fault and 2 cannot be compared.

\begin{table}[t]
\caption{\textbf{First revisions of the same 417 failed first answers} (\terra, three runs pooled). Fresh retries see no feedback; binary feedback says only that the answer failed.}
\label{tab:revisions}
\centering
\tablestyle\setlength{\tabcolsep}{3.4pt}
\begin{tabular}{@{}lccc@{}}
\toprule
& Fresh & Binary & Visual \\
\midrule
Revision solves the puzzle & 33 & 39 & 56 \\
Every earlier fault cleared & 70 & 111 & 175 \\
\quad and the puzzle solved & 18 & 24 & 43 \\
\quad but failing elsewhere & 52 & 87 & 132 \\
Car no longer reaches the fault & 44 & 80 & 71 \\
Same failure repeated & 79 & 68 & 20 \\
Fails before reaching the fault & 70 & 42 & 28 \\
No comparison possible & 154 & 116 & 123 \\
Same layout as the failed one & 59 & 25 & 8 \\
Changed cells (median) & 8 & 8 & 7 \\
Changed cells near the fault (\%) & 34.7 & 37.0 & 46.8 \\
Solved later, within five answers & 74 & 72 & 91 \\
\bottomrule
\end{tabular}
\end{table}

\section{Prompts and feedback messages}
\label{sup:prompts}

The first prompt contains, in order: the task instruction; the board size and coordinate convention; what may and may not change; the track budget; the names of the rail and mechanism codes that occur on this board; the answer format; how pieces are drawn in the image; the shared movement rules; the goal; a lookup table of how each rail piece turns a car; and the rule of every mechanism present (\cref{fig:prompt}).
The file \texttt{prompts.py} of the released code produces the prompt of every puzzle, and \cref{fig:prompt} reproduces its output for one puzzle.
With text input, the description of the drawing is replaced by the rail grid as numbers, a list of mechanisms by position, and the list of cars.
After a failure the model receives one of the messages in \cref{fig:feedback-text}; an answer over budget receives the budget message in every condition.
Each later answer must again list all new rails relative to the original board.

\input{sec/supp_prompt}

\section{Related evaluation settings}
\label{sup:related}

\Cref{tab:compare} compares the evaluation setting of \bench with recent benchmarks discussed in \mainref{sec:related}{Sec.~2 of the main paper}.

\paragraph{Related findings.}
Several results elsewhere bear on ours.
Repeated sampling raises the share of problems that some sample solves~\cite{brown2024monkeys}; the fresh-retry policy measures this effect on our puzzles, so feedback is always compared with it.
For reasoning and planning tasks, self-critique by the model itself is unreliable, and most of the gain comes from a sound external verifier~\cite{stechly2025selfverification}; our simulator is such a verifier, and we vary what it reports back.
Language models also struggle to find collision-free paths for several agents on a grid~\cite{chen2024mapf}; in \bench the routes must also share rails, wait at mechanisms, and arrive in order.
Logic and algorithmic puzzle benchmarks find that accuracy drops as puzzles grow harder~\cite{lin2025zebralogic,fan2024nphardeval} and that multimodal models do poorly on visual algorithmic puzzles~\cite{ghosal2025algopuzzlevqa}, and studies of perception and spatial reasoning find that these models often miss simple visual details or gain little from an image over a text description~\cite{fu2024blink,rahmanzadehgervi2024blind,wang2024spatialeval}.
Our copying and prediction tests separate these two sources of error from the construction of a plan.

\begin{table}[t]
\caption{\textbf{Related evaluation settings.} \bench judges first plans and revised plans with the same simulator, and varies the information and feedback on fixed puzzles.}
\label{tab:compare}
\centering
\tablestyle
\begin{tabular}{@{}lll@{}}
\toprule
Benchmark & Model response & When revision happens \\
\midrule
USTBench~\cite{lai2026ustbench} & urban decisions, QA & reflection on feedback \\
SpaceAct~\cite{shi2026spaceact} & physical actions & closed-loop correction \\
AsgardBench~\cite{tupini2026asgardbench} & remaining plan & during execution \\
BuildArena~\cite{xia2026buildarena} & structure design & across trials \\
\bench & complete layout & after a complete run \\
\bottomrule
\end{tabular}
\end{table}

\section{Data and records}
\label{sup:data}

\paragraph{Puzzle inputs.}
Every leaderboard and diagnostic run stores the puzzle inputs it used, field by field (board, mechanisms, mechanism numbers, cars, track budget and signal budget), and these are identical to the released puzzle files.

\paragraph{Code and data.}
\ifarxiv
The puzzles, reference solutions, simulator, solver and evaluation code are open-sourced on GitHub at \codeurl.
\else
The archive submitted with this supplement contains the three puzzle sets (with the 233 original levels), their reference solutions, the simulator, the solver, the evaluation code, and the scripts for the duplicate check and the random-layout baseline of \cref{sup:puzzles}; it does not contain run records.
\fi

\paragraph{Records.}
For every graded answer the records keep the prompt sources, the raw reply, the judged outcome with its failure type and fault cells, the feedback that followed, and the reported token usage.
The statistics in this supplement are recomputed from the released puzzle and solution files and these records.

%% file: sec/supp_prompt.tex
\begin{figure*}[p]
\centering
\fbox{\parbox{0.97\textwidth}{\raggedright\ttfamily\fontsize{6.0pt}{7.0pt}\selectfont
Solve this railway-routing puzzle using your own reasoning only. You have no tools,\allowbreak{} code execution,\allowbreak{} browsing,\allowbreak{} agents,\allowbreak{} or reference solutions. Return exactly one JSON object,\allowbreak{} without Markdown fences or commentary. Do not request or simulate tool calls.
\par\smallskip
Solve level "E-CT0013". The board has 4 rows and 5 columns. Coordinates are zero-based [row,\allowbreak{}column]\allowbreak{},\allowbreak{} row 0 at the top and column 0 at the left. Valid rows are 0 through 3 and columns 0 through 4. Never add cells outside this board.
\par\smallskip
Preserve every original nonzero board tile,\allowbreak{} every original nonzero modifier and numbered mechanism ID. Initial cars,\allowbreak{} directions,\allowbreak{} types,\allowbreak{} numbers,\allowbreak{} dimensions and budgets are fixed. Add rails only to empty cells.
\par\smallskip
Maximum number of new track cells:\allowbreak{} 8. Each added straight,\allowbreak{} curved or three-way rail cell costs one track. This is an upper bound,\allowbreak{} not an exact requirement; unused tracks are allowed. Original rails do not consume this budget.
\par\smallskip
Tile code names:\allowbreak{} \{"0":\allowbreak{}"empty",\allowbreak{}"1":\allowbreak{}"horizontal",\allowbreak{}"2":\allowbreak{}"vertical",\allowbreak{}"4":\allowbreak{}"obstacle",\allowbreak{}"5":\allowbreak{}"bottom\_right",\allowbreak{}"6":\allowbreak{}"bottom\_left",\allowbreak{}"7":\allowbreak{}"top\_right",\allowbreak{}"8":\allowbreak{}"top\_left",\allowbreak{}"9":\allowbreak{}"bottom\_right\_left",\allowbreak{}"10":\allowbreak{}"bottom\_right\_top",\allowbreak{}"11":\allowbreak{}"bottom\_left\_right",\allowbreak{}"12":\allowbreak{}"bottom\_left\_top",\allowbreak{}"13":\allowbreak{}"top\_right\_left",\allowbreak{}"14":\allowbreak{}"top\_right\_bottom",\allowbreak{}"15":\allowbreak{}"top\_left\_right",\allowbreak{}"16":\allowbreak{}"top\_left\_bottom",\allowbreak{}"29":\allowbreak{}"passenger\_end\_left"\}
\par\smallskip
Modifier codes:\allowbreak{} \{"0":\allowbreak{}"none",\allowbreak{}"6":\allowbreak{}"passenger\_station",\allowbreak{}"7":\allowbreak{}"self\_switching\_junction",\allowbreak{}"10":\allowbreak{}"car\_start"\}
\par\smallskip
Output format:\allowbreak{} exactly one JSON object with required key add\_tracks,\allowbreak{} an array of [row,\allowbreak{}column,\allowbreak{}track\_code]\allowbreak{} entries for NEW rail cells only. Example syntax only:\allowbreak{} \{"add\_tracks":\allowbreak{}[[0,\allowbreak{}0,\allowbreak{}1]\allowbreak{}]\allowbreak{}\}. Coordinates are zero-based [row,\allowbreak{}column]\allowbreak{} within the original board. Use JSON integers,\allowbreak{} never booleans. Do not list original rails,\allowbreak{} cars,\allowbreak{} dimensions,\allowbreak{} budgets,\allowbreak{} metadata or mechanism IDs. Everything not listed is inherited unchanged from the ORIGINAL level. Each round return your complete proposed additions relative to that original level,\allowbreak{} not an incremental edit to your previous answer. No duplicate positions,\allowbreak{} extra keys,\allowbreak{} comments,\allowbreak{} Markdown fences or commentary.
\par\smallskip
All modifiers and numbered mechanism IDs remain unchanged.
\par\smallskip
Read the complete starting board from the image. Column coordinates are printed above the grid; row coordinates are on the left. Every grid square is one cell. Empty terrain is board code 0; a rock is code 4. Steel-colored rails are original; amber rails in later feedback are additions. For a straight/curved rail,\allowbreak{} its drawn endpoints determine its board code. For a three-way rail,\allowbreak{} the solid curved rail gives the first two endpoints in its code name,\allowbreak{} and the third word gives its other endpoint. In code names,\allowbreak{} top=UP and bottom=DOWN. This description specifies the drawing and numeric encoding.\\
Vehicles have a blunt rear,\allowbreak{} a pointed front and a white arrow on the nose. The arrow determines direction (LEFT,\allowbreak{} RIGHT,\allowbreak{} UP,\allowbreak{} DOWN). Blue passenger cars use Arabic labels 1,\allowbreak{} 2,\allowbreak{} ... and type NORMAL; their num is the visible label minus 1. A blue locomotive has a cab,\allowbreak{} boiler and chimney,\allowbreak{} and its board code passenger\_end\_DIRECTION uses its arrow direction. A locomotive is a board tile,\allowbreak{} never an entry in cars. A car may share its starting cell with a drawn mechanism. Modifier 10 marks a plain car start; it does not replace a mechanism shown in that cell. The rail visible around the vehicle still determines that cell's board code.\\
A physical points lever joined to the rail tongue,\allowbreak{} with curved arrows on an unnumbered green handle,\allowbreak{} is modifier 7 (self\_switching\_junction). Its current bend and straight rail are solid and have identical colors. The inactive bend is drawn as dashed rails; use only the solid curved rail to encode the selected pair.\\
A cream platform beside the rail,\allowbreak{} with boarding steps,\allowbreak{} a waving person carrying a suitcase and an Arabic number sign,\allowbreak{} is modifier 6 (passenger\_station). It belongs to the rail cell containing the platform; mod\_nums is its label minus 1.\\
Unmarked cells have modifier 0. Unnumbered modifiers and cells without a printed ID use mod\_nums 0. A served platform in feedback is empty,\allowbreak{} with a green check in place of its person or parcel; consumed modifiers use code 9. Red outlines and exclamation badges mark feedback error cells. In a failed frame,\allowbreak{} translucent vehicles show previous positions; solid vehicles depict the failure instant. Red motion trails show the last attempted movement,\allowbreak{} and white/orange bursts show contact points. These overlays are not additional vehicles,\allowbreak{} rails or modifiers.
\par\smallskip
Infer all original cells and vehicle positions,\allowbreak{} identities and directions from the image. No serialized original board is supplied.
\par\smallskip
All cars move simultaneously by one cell each tick unless waiting. A car direction is its next movement direction. Reach the matching locomotive in number order,\allowbreak{} starting with num 0. Delivery removes that car at the end of its move,\allowbreak{} with no extra global coupling pause; other cars continue. Two cars occupying the same cell,\allowbreak{} exchanging adjacent cells head-on,\allowbreak{} or entering invalid/missing rail crash. Following a moving car into its vacated cell is allowed. Numbered delivery order is global for each car type across all of its matching locomotives. All required cars must be delivered. Entering a loop fails.
\par\smallskip
Goal:\allowbreak{} before delivery,\allowbreak{} every required car must visit every stop assigned to its type and zero-based number. Passenger stations belong to NORMAL cars; post offices belong to NUMERAL cars.
\par\smallskip
Explicit rail movement lookup. Incoming is the car movement direction BEFORE entering the tile,\allowbreak{} not its entry side:\allowbreak{} moving RIGHT enters from the LEFT. Outgoing is the heading selected on entry and used for the NEXT move. Missing incoming directions are incompatible. top=UP; bottom=DOWN.\\
1:\allowbreak{} LEFT -\textgreater{} LEFT,\allowbreak{} RIGHT -\textgreater{} RIGHT\\
2:\allowbreak{} DOWN -\textgreater{} DOWN,\allowbreak{} UP -\textgreater{} UP\\
5:\allowbreak{} LEFT -\textgreater{} DOWN,\allowbreak{} UP -\textgreater{} RIGHT\\
6:\allowbreak{} RIGHT -\textgreater{} DOWN,\allowbreak{} UP -\textgreater{} LEFT\\
7:\allowbreak{} LEFT -\textgreater{} UP,\allowbreak{} DOWN -\textgreater{} RIGHT\\
8:\allowbreak{} RIGHT -\textgreater{} UP,\allowbreak{} DOWN -\textgreater{} LEFT\\
9:\allowbreak{} LEFT -\textgreater{} DOWN,\allowbreak{} RIGHT -\textgreater{} RIGHT,\allowbreak{} UP -\textgreater{} RIGHT\\
10:\allowbreak{} LEFT -\textgreater{} DOWN,\allowbreak{} DOWN -\textgreater{} DOWN,\allowbreak{} UP -\textgreater{} RIGHT\\
11:\allowbreak{} LEFT -\textgreater{} LEFT,\allowbreak{} RIGHT -\textgreater{} DOWN,\allowbreak{} UP -\textgreater{} LEFT\\
12:\allowbreak{} RIGHT -\textgreater{} DOWN,\allowbreak{} DOWN -\textgreater{} DOWN,\allowbreak{} UP -\textgreater{} LEFT\\
13:\allowbreak{} LEFT -\textgreater{} UP,\allowbreak{} RIGHT -\textgreater{} RIGHT,\allowbreak{} DOWN -\textgreater{} RIGHT\\
14:\allowbreak{} LEFT -\textgreater{} UP,\allowbreak{} DOWN -\textgreater{} RIGHT,\allowbreak{} UP -\textgreater{} UP\\
15:\allowbreak{} LEFT -\textgreater{} LEFT,\allowbreak{} RIGHT -\textgreater{} UP,\allowbreak{} DOWN -\textgreater{} LEFT\\
16:\allowbreak{} RIGHT -\textgreater{} UP,\allowbreak{} DOWN -\textgreater{} LEFT,\allowbreak{} UP -\textgreater{} UP
\par\smallskip
Locomotive entry requirements:\allowbreak{} \{"29":\allowbreak{}\{"car\_type":\allowbreak{}"NORMAL",\allowbreak{}"incoming\_direction":\allowbreak{}"LEFT"\}\}. A RIGHT-facing locomotive is entered by moving RIGHT from its LEFT neighbor.
\par\smallskip
A switching event replaces a junction code using \{"9":\allowbreak{}11,\allowbreak{}"11":\allowbreak{}9,\allowbreak{}"10":\allowbreak{}14,\allowbreak{}"14":\allowbreak{}10,\allowbreak{}"12":\allowbreak{}16,\allowbreak{}"16":\allowbreak{}12,\allowbreak{}"13":\allowbreak{}15,\allowbreak{}"15":\allowbreak{}13\}.
\par\smallskip
Junction rules:\allowbreak{} A three-way rail has a preferred curved pair and a straight branch. The first two direction words in its name are its curved endpoints; the third is the extra straight endpoint. Entering a curved endpoint follows the curve; entering the extra branch continues straight. Ordinary junctions keep their orientation. Any newly occupied rail cell costs one track,\allowbreak{} including a junction.
\par\smallskip
Automatic junction rules:\allowbreak{} A self-switching junction changes its preferred curve after a new arrival has selected its outgoing heading,\allowbreak{} before the next movement opportunity,\allowbreak{} with zero extra waiting ticks. A waiting car does not repeatedly switch it.
\par\smallskip
Passenger station rules:\allowbreak{} Each station ID is assigned to exactly the NORMAL car with that zero-based number. Other cars pass without collecting it. On movement entry,\allowbreak{} its matching car stops for TWO additional movement opportunities,\allowbreak{} then moves on. Each passenger station must be served once before its car reaches its locomotive. Multiple stations may belong to one car.
}}
\caption{\textbf{The first prompt for puzzle E-CT0013}, the puzzle of \mainref{fig:teaser}{Fig.~1}. Line breaks within paragraphs are added for typesetting. When the mechanism rules are withheld, the movement lookup and the mechanism paragraphs are replaced by: \emph{Mechanism transition rules and waiting durations are not supplied. Use the shared goal,\allowbreak{} visual symbols,\allowbreak{} output format and observed execution feedback.}}
\label{fig:prompt}
\end{figure*}

\begin{figure*}[t]
\centering\small
\begin{tabular}{@{}p{0.17\textwidth}p{0.78\textwidth}@{}}
\toprule
Binary & {\raggedright\ttfamily\fontsize{5.6pt}{6.6pt}\selectfont The solution failed. Please submit another complete answer.} \\
Error type and cells & {\raggedright\ttfamily\fontsize{5.6pt}{6.6pt}\selectfont Failure:\allowbreak{} derailment. Affected [row,\allowbreak{}column]\allowbreak{} cells:\allowbreak{} [[3,\allowbreak{} 3]\allowbreak{},\allowbreak{} [3,\allowbreak{} 4]\allowbreak{}]\allowbreak{}. Submit another complete additions JSON.} \\
Text states & {\raggedright\ttfamily\fontsize{5.6pt}{6.6pt}\selectfont Failure:\allowbreak{} derailment. Affected [row,\allowbreak{}column]\allowbreak{} cells:\allowbreak{} [[3,\allowbreak{} 3]\allowbreak{},\allowbreak{} [3,\allowbreak{} 4]\allowbreak{}]\allowbreak{}. Selected feedback observations (chronological):\allowbreak{} [\{"tick":\allowbreak{}0,\allowbreak{}"board":\allowbreak{}[[29,\allowbreak{}9,\allowbreak{}1,\allowbreak{}9,\allowbreak{}6]\allowbreak{},\allowbreak{}[0,\allowbreak{}2,\allowbreak{}0,\allowbreak{}10,\allowbreak{}16]\allowbreak{},\allowbreak{}[0,\allowbreak{}7,\allowbreak{}6,\allowbreak{}2,\allowbreak{}2]\allowbreak{},\allowbreak{}[0,\allowbreak{}0,\allowbreak{}7,\allowbreak{}13,\allowbreak{}0]\allowbreak{}]\allowbreak{},\allowbreak{}"delivered":\allowbreak{}\{"NORMAL":\allowbreak{}[]\allowbreak{},\allowbreak{}"NUMERAL":\allowbreak{}[]\allowbreak{}\},\allowbreak{}"mechanisms":\allowbreak{}[\{"pos":\allowbreak{}[0,\allowbreak{}1]\allowbreak{},\allowbreak{}"modifier":\allowbreak{}7,\allowbreak{}"id":\allowbreak{}0\},\allowbreak{}\{"pos":\allowbreak{}[0,\allowbreak{}3]\allowbreak{},\allowbreak{}"modifier":\allowbreak{}7,\allowbreak{}"id":\allowbreak{}0\},\allowbreak{}\{"pos":\allowbreak{}[1,\allowbreak{}1]\allowbreak{},\allowbreak{}"modifier":\allowbreak{}6,\allowbreak{}"id":\allowbreak{}0\},\allowbreak{}\{"pos":\allowbreak{}[2,\allowbreak{}4]\allowbreak{},\allowbreak{}"modifier":\allowbreak{}6,\allowbreak{}"id":\allowbreak{}1\}]\allowbreak{},\allowbreak{}"cars":\allowbreak{}[\{"id":\allowbreak{}0,\allowbreak{}"type":\allowbreak{}"NORMAL",\allowbreak{}"num":\allowbreak{}1,\allowbreak{}"pos":\allowbreak{}[0,\allowbreak{}2]\allowbreak{},\allowbreak{}"direction":\allowbreak{}"RIGHT"\},\allowbreak{}\{"id":\allowbreak{}1,\allowbreak{}"type":\allowbreak{}"NORMAL",\allowbreak{}"num":\allowbreak{}0,\allowbreak{}"pos":\allowbreak{}[2,\allowbreak{}3]\allowbreak{},\allowbreak{}"direction":\allowbreak{}"DOWN"\}]\allowbreak{}\},\allowbreak{} ... ]\allowbreak{}} \\
Visual frames & {\raggedright\ttfamily\fontsize{5.6pt}{6.6pt}\selectfont Failure:\allowbreak{} derailment. Affected [row,\allowbreak{}column]\allowbreak{} cells:\allowbreak{} [[3,\allowbreak{} 3]\allowbreak{},\allowbreak{} [3,\allowbreak{} 4]\allowbreak{}]\allowbreak{}. The attached feedback images are in chronological order at ticks [0,\allowbreak{} 2]\allowbreak{}. Faded vehicles show the preceding pose,\allowbreak{} red trails show the attempted movement,\allowbreak{} and impact bursts mark contact. An unserved stop is highlighted at its platform. For a loop,\allowbreak{} the last image shows where the loop begins. Submit another complete additions JSON.} \\
\bottomrule
\end{tabular}
\caption{\textbf{Feedback messages after one failed layout} of puzzle E-CT0013 (the reference layout without its last added rail; a car derails at step 2). The text-state message lists the selected states in full; only the first is shown. Visual feedback attaches the rendered frames to the message.}
\label{fig:feedback-text}
\end{figure*}